\documentclass{article}

\usepackage{PRIMEarxiv}
\usepackage{amsmath}
\usepackage[utf8]{inputenc} % allow utf-8 input
\usepackage[T1]{fontenc}    % use 8-bit T1 fonts
\usepackage{hyperref}       % hyperlinks
\usepackage{url}            % simple URL typesetting
\usepackage{booktabs}       % professional-quality tables
\usepackage{amsfonts}       % blackboard math symbols
\usepackage{nicefrac}       % compact symbols for 1/2, etc.
\usepackage{microtype}      % microtypography
\usepackage{lipsum}
\usepackage{fancyhdr}       % header
\usepackage{graphicx}       % graphics
\graphicspath{{media/}}     % organize your images and other figures under media/ folder
\usepackage{xcolor}
\usepackage{xspace}
\usepackage{authblk}
\usepackage{float}
\usepackage{enumitem}
\usepackage{tabularx}
\usepackage{listings}
\newcommand{\modelname}{\texttt{o1}\xspace{ }}

\title{Towards Next-Generation Medical Agent: How \modelname is Reshaping Decision-Making in Medical Scenarios}

\author[1]{Shaochen Xu$^{*}$}
\author[1]{Yifan Zhou$^{*}$}
\author[1]{Zhengliang Liu}
\author[1]{Zihao Wu}
\author[2]{Tianyang Zhong}
\author[1]{Huaqin Zhao}
\author[1]{Yiwei Li}
\author[1]{Hanqi Jiang}
\author[1]{Yi Pan}
\author[1]{Junhao Chen}
\author[1]{Jin Lu}
\author[3]{Wei Zhang}
\author[2]{Tuo Zhang}
\author[4]{Lu Zhang}
\author[5]{Dajiang Zhu}
\author[6]{Xiang Li}
\author[7]{Wei Liu}
\author[6]{Quanzheng Li}
\author[8]{Andrea Sikora}
\author[9,10,11]{Xiaoming Zhai}
\author[1]{Zhen Xiang$^{\dagger}$}
\author[1,10,11]{Tianming Liu$^{\dagger}$}

\affil[1]{School of Computing, University of Georgia, Athens, GA, USA}
\affil[2]{Department of Mathematical and Statistical Sciences, University of Alberta, Edmonton,
Canada}
\affil[3]{School of Computer and Cyber Sciences, Augusta University, Augusta, GA, USA}
\affil[4]{Department of Computer Science, Indiana University, Indianapolis, IN,
USA}
\affil[5]{Department of Computer Science and Engineering, University of Texas, Arlington, TX,
USA}
\affil[6]{Department of Radiology, Massachusetts General Hospital and Harvard Medical School, Boston, MA,
USA}
\affil[7]{Department of Radiation Oncology, Mayo Clinic, Scottsdale, AZ, USA}
\affil[8]{Department of Biomedical Informatics, University of Colorado, Aurora, CO, USA}
\affil[9]{Department of Mathematics, Science, and Social Studies Education, University of Georgia, Athens, GA, USA}
\affil[10]{AI4STEM Education Center, University of Georgia, Athens, GA, USA}
\affil[11]{National GENIUS Center, University of Georgia, Athens, GA, USA}

\begin{document}
\maketitle
\footnotetext[1]{\textsuperscript{*}Equal Contribution.}
\footnotetext[2]{\textsuperscript{†}Joint Correspondence Authors, emails: zxiangaa@uga.edu and tliu@uga.edu}

\begin{abstract}

Artificial Intelligence (AI) has become essential in modern healthcare, with large language models (LLMs) offering promising advances in clinical decision-making.
Traditional model-based approaches, including those leveraging in-context demonstrations and those with specialized medical fine-tuning, have demonstrated strong performance in medical language processing but struggle with real-time adaptability, multi-step reasoning, and handling complex medical tasks.
Agent-based AI systems address these limitations by incorporating reasoning traces, tool selection based on context, knowledge retrieval, and both short- and long-term memory.
These additional features enable the medical AI agent to handle complex medical scenarios where decision-making should be built on real-time interaction with the environment.
Therefore, unlike conventional model-based approaches that treat medical queries as isolated questions, medical AI agents approach them as complex tasks and behave more like human doctors.
In this paper, we study the choice of the backbone LLM for medical AI agents, which is the foundation for the agent’s overall reasoning and action generation.
In particular, we consider the emergent o1 model and examine its impact on agents' reasoning, tool-use adaptability, and real-time information retrieval across diverse clinical scenarios, including high-stakes settings such as intensive care units (ICUs).
Our findings demonstrate o1’s ability to enhance diagnostic accuracy and consistency, paving the way for smarter, more responsive AI tools that support better patient outcomes and decision-making efficacy in clinical practice.

\end{abstract}

\section{Introduction}
The medical environment requires dynamic artificial intelligence (AI) tools. Foundational large language models (LLMs) like GPT-3.5 and GPT-4 have transformed how the medical field interacts with information technology (IT). Beyond serving as powerful tools for knowledge synthesis \cite{zhong2024evaluation,liu2023holistic}, general-purpose LLMs like GPT-4 can achieve excellent accuracy handling domain-specific queries when provided with in-context demonstrations of medical reasoning -- they serve as reasoning engines to support complex decision-making.~\cite{nori2023capabilitiesgpt4medicalchallenge,holmes2023evaluating,li2024echopulse}. Moreover, specialized medical models like MedPaLM2 and MedGemini have been trained or fine-tuned on extensive medical datasets and medical textbooks to install medicine-specific reasoning capabilities.  \cite{nori2023generalistfoundationmodelsoutcompete}~\cite{saab2024capabilitiesgeminimodelsmedicine} 
These specialized LLMs have excelled in comprehensive evaluations with state-of-the-art benchmarks, highlighting the significant progress that has been made with AI healthcare. However, both foundational and fine-tuned medical models face limitations that hinder effectiveness in real-world medical applications.~\cite{li2023artificial}.

To date, LLMs excel at static tasks (e.g., answering questions based on pre-existing data), but they lack the capability to interact dynamically with the ever-changing medical environment \cite{nori2023capabilitiesgpt4medicalchallenge, liévin2023largelanguagemodelsreason}. These limitations show up in two important limitations: (1) rapid updates in medical knowledge and (2) rapid changes in individual patient conditions.
First, the corpus of medical knowledge is constantly changing.
Paradigm-shifting studies have been presented at conferences that change the standard of care overnight, despite that guidelines may lag years behind in updates.  Indeed, the pace of medical knowledge doubling has shortened from 50 years to just 73 days \cite{densen2011challenges}. Second, critically ill patients have conditions that require rapid assimilation of multi-modal data for assessment and decisions that must occur within minutes to provide life-saving care. Multi-modal data extraction from vital sign flowsheets, laboratory values, medication administration records, and imaging are essential bedside capabilities \cite{10.1093/jamia/ocac064}. 

Unlike foundational models, LLM agents are characterized by integrated capabilities to interact with a dynamic environment, conduct knowledge-assisted reasoning, and take appropriate actions based on the user’s queries   ~\cite{shi2024ehragent, abbasian2024conversationalhealthagentspersonalized, tu2024conversationaldiagnosticai, yang2024psychogat, dai2023ad, xiao2024comprehensivesurveylargelanguage}.
LLM agents are equipped with a knowledge base for the retrieval of task-related information and a toolbox of local functions or third-party APIs s~\cite{parisi2022talmtoolaugmentedlanguage, schick2023toolformer}. These agents use one or more LLMs as the backbone for reasoning and subsequent action (e.g., decision for knowledge retrieval or selection of the most appropriate tool). The extendable knowledge base and the adaptive toolbox, integrated with the LLM-based reasoning engine, enable LLM agents to handle complex tasks in diverse application domains, including the medical scenarios discussed herein. ~\cite{yu2023finmemperformanceenhancedllmtrading, li2024agenthospitalsimulacrumhospital, deng2023mind2webgeneralistagentweb, xiang2024guardagent, gur2024a, chen2024agentpoisonredteamingllmagents, li2024idsagent}

Here, we examine the impact of the backbone LLM choice on medical agents, with a special focus on how the emergent o1 model affects agent performance. o1 model is distinguished with a Chain-of-Thought (CoT) reasoning framework embedded directly into its training process combined with reinforcement learning from human feedback (RLHF) to enhance its decision-making capabilities  \cite{wei2023chainofthoughtpromptingelicitsreasoning}. Recent evaluations highlight that o1 outperforms its predecessors on a variety of medical datasets, showing particular strength in tasks that require multi-step reasoning, such as those found in the LancetQA and NEJMQA datasets\cite{xie2024preliminarystudyo1medicine}. These findings suggest that o1 could play a major role in refining the performance of AI agents in medicine, where reasoning and the ability to interact with the external environment and tool use are essential. Furthermore, o1’s capability to integrate retrieval-augmented generation (RAG) techniques enables it to access and use up-to-date information, making it more adaptable and relevant in clinical settings where current medical knowledge is essential. We explore the practical benefits of integrating o1 into medical AI agents as their backbone by considering the following three agents:

\textbf{Chain of Diagnosis} (CoD) \cite{chen2024chain} is a medical diagnostics framework designed with a particular focus on its interpretability. The pipeline of CoD mirrors the cognitive workflows of healthcare professionals, including symptom abstraction, disease retrieval, and diagnostic reasoning.
\textbf{MedAgents} \cite{tang2023medagents} is a framework that facilitates collaborative medical decision-making through the integration of multiple domain experts via ``role-play''. MedAgents is characterized by a structured workflow that selects a group of different types of medical specialists, gathers insights from these specialists, aggregates their inputs, and then refines the medical decisions iteratively.
\textbf{AgentClinic} serves as a sophisticated framework of agents that simulates real-world clinical environments to evaluate AI-driven language models. It features a different multi-agent pipeline where a patient agent interacts with a doctor agent and a third measurement agent provides clinical test results. ~\cite{schmidgall2024agentclinicmultimodalagentbenchmark}

We tested these three agents using o1 as the backbone, compared with the previous state-of-the-art backbone model GPT-4, on various popular datasets including MedQA, DxBench, and NEJMQA. We also replaced GPT-4 with o1 for each module of the three agents to further investigate the potential enhancements by o1 in agent diagnostic accuracy, reasoning consistency, and reliability. The key findings from our experiments reveal several improvements with the integration of o1:

\begin{itemize}
     \item {Enhanced Reasoning Framework}: The o1 model demonstrated significant improvements in diagnostic accuracy across clinical benchmarks like DxBench and NEJMQA, largely due to its enhanced reasoning framework. By leveraging advanced techniques such as Chain-of-Thought (CoT) prompting and retrieval-augmented generation (RAG), o1 effectively handles complex diagnostic queries and simulates clinical workflows, particularly excelling in multi-step reasoning tasks.
     \item {Reduction in Diagnostic Variability:} Across tasks requiring decision-making, o1 exhibited more consistent performance, with reduced variability in accuracy compared to prior models. This suggests a potential for greater reliability in diagnostic tasks across diverse scenarios.
     \item {Possible Drawback of o1:} While o1 excels at complex reasoning tasks, it tends to underperform in simpler tasks that do not require advanced reasoning, such as expert gathering in MedAgents or information provision by the patient agent in AgentClinic. Additionally, its increased computational demands make it less efficient for these straightforward operations, where simpler models like GPT-4 may suffice.
\end{itemize}

Together, these findings convey the advantages of o1 in elevating diagnostic performance and reliability across varied medical scenarios, pointing to its value in high-stakes clinical applications. While also revealing its limitations that can be improved in future work. By building on the preliminary findings that highlight o1’s superior performance, this study seeks to demonstrate how these improvements translate into real-world healthcare settings. Ultimately, by harnessing the advanced reasoning and adaptability of o1, we can move closer to developing AI agents that meet the demanding needs of modern medicine.

\section{Related Work}

\subsection{LLMs in Medical Domains}
In recent years, significant progress has been made in the application of LLMs within the medical field \cite{li2023artificial,liu2023evaluating,liu2023holistic}. These advancements have seen LLMs being applied in various medical domains such as computer-aided diagnosis, genotyping and phenotyping, drug discovery, medical evidence summarization, and treatment planning in radiation oncology \cite{zhang2024generalist,liu2023radiology,dai2023ad,wu2023exploring,liu2023pharmacygpt,ma2024iterative,lyu2024gp}. However, despite their potential, LLMs still face challenges when confronted with clinical tasks that require complex reasoning and specialized medical domain knowledge \cite{zhong2024evaluation,zhao2023brain}, both of which are essential for enhancing LLMs' effectiveness across clinical scenarios.

Currently, two major research directions exist in this area: tool-augmented methods \cite{lewis2020retrieval} and instruction-tuning methods \cite{alpaca}. In the tool-augmented approach, LLMs are equipped with external tools to gather additional medical information for better reasoning. For example, GeneGPT \cite{jin2024genegpt} integrates Web APIs from the National Center for Biotechnology Information (NCBI) to meet biomedical information needs. Similarly, Almanac \cite{zakka2024almanac} integrates a RAG framework with GPT-4 designed for accessing medical guidelines and treatment recommendations. Another notable method, KARD \cite{kang2024knowledge}, enhances domain-specific knowledge by fine-tuning smaller language models with rationales from LLMs and augmenting them with external knowledge from non-parametric memory sources.

On the other hand, instruction-tuning research leverages external clinical knowledge bases and synthetic datasets to create specialized instruction datasets. These datasets are used to fine-tune LLMs for specific medical applications. Some models are trained on a vast array of medical and biomedical literature to gain broad medical insights \cite{li2023artificial,li2024llava}, while others focus on specific fields such as radiology report generation for task-oriented design \cite{liu2023radiology,lyu2024gp}. In contrast to these methods, our work aims to leverage the latent medical knowledge within LLMs and improve clinical reasoning in a training-free setting.

In addition to the existing literature, recent studies like the one by Xie et al. \cite{xie2024preliminarystudyo1medicine} have provided a broad evaluation of advanced LLMs, such as o1, specifically examining their diagnostic and reasoning capabilities across diverse medical scenarios. While these studies highlight the potential of LLMs in medicine by focusing on understanding, reasoning, and handling clinical data, our study emphasizes the performance of LLMs within the context of agent systems. Here, each agent must not only apply medical knowledge but also engage dynamically in knowledge retrieval and reasoning-based tool usage. Thus, while standalone LLMs demonstrate impressive capabilities, integrating them into agent-based frameworks requires models adept at collaborative decision-making.

\subsection{LLM Agents and Their Applications in the Medical Domain}
Recent advancements in LLM agents have enabled these systems to act autonomously in complex environments by leveraging advanced reasoning capabilities and memory-based reflection. These agents can dynamically respond to new information, process complex inputs, and make decisions based on their understanding of context \cite{yao2023react, shinn2023reflexion}. Moreover, by utilizing memory, they can maintain continuity across multi-turn interactions, adapting their responses in line with previous exchanges—much like a human professional would in an ongoing consultation~\cite{zhang2024surveymemorymechanismlarge}.

A critical feature of effective LLM agents is their ability to reason through complex tasks using multi-step processes. Approaches like \emph{Chain-of-Thought} (CoT) prompting have been developed to enhance the reasoning depth of LLMs, allowing them to break down problems into sequential steps that mirror human logical reasoning \cite{wei2022chain}. This method has proven especially valuable in the medical field, where diagnostic reasoning requires careful consideration of patient symptoms, history, and test results. By generating intermediate steps, CoT prompting enables LLM agents to better navigate the complexities inherent in medical decision-making.

In the medical domain, specialized LLM agents like Med-PaLM \cite{singhal2023large} and BioGPT \cite{luo2022biogpt} have emerged to address the unique demands of healthcare settings. These agents are equipped with capabilities such as advanced knowledge retrieval, adherence to clinical guidelines, and robust multi-step reasoning to support diagnostic and therapeutic decisions. Training on large-scale medical datasets further equips these models to interpret clinical notes, answer complex medical queries, and synthesize patient data effectively \cite{chang2023learning}. Reinforcement Learning from Human Feedback (RLHF) plays a critical role here, aligning the model’s outputs with human expertise and judgment. The o1 model, for instance, leverages both CoT and RLHF, allowing it to produce reliable reasoning outcomes—a quality essential for high-stakes clinical applications \cite{zhong2024evaluationopenaio1opportunities}.

Beyond individual LLM agents, recent research has explored multi-agent systems within the medical domain, where multiple LLM agents work together to provide more comprehensive clinical insights. These systems are inspired by the collaborative nature of healthcare, where doctors, specialists, and other healthcare providers work in concert to address complex cases. Multi-agent frameworks aim to replicate this teamwork by assigning specific roles to LLM agents, such as patient, doctor, and other necessary roles, which enables them to simulate consultations and cooperative diagnostics \cite{kleinpell2021, ervin2018}. By drawing on role-playing techniques from frameworks like Camel, these agents engage in interactive discussions and debates, refining their diagnoses through multi-turn feedback \cite{chen2023universal}. This collaborative approach holds promise for emulating the multidisciplinary decision-making processes crucial in healthcare, ultimately advancing the quality and depth of AI-assisted patient care.

Given these advancements, our study leverages agent frameworks with robust reasoning to address the complexities of medical applications. These frameworks, designed to support both independent and cooperative decision-making, reflect real-world healthcare dynamics, where diagnoses and treatment plans often involve multidisciplinary input. By incorporating reasoning, memory-based reflection, collaborative feedback, etc., our chosen frameworks aim to enhance diagnostic accuracy and align closely with the collaborative nature of clinical practice.

\section{Agent}
\subsection{CoD Agent}

\textbf{Overview:} The Chain of Diagnosis (CoD) \cite{chen2024chain} framework is a medical AI agent designed for automated diagnosis.
It is characterized by a pipeline that simulates the cognitive workflows of healthcare professionals. When a patient is seen in a healthcare setting, the healthcare professional goes through a systematic process summarized by the SOAP format (subjective, objective, assessment, plan) \cite{podder2023soap}, which includes collecting subjective symptoms (e.g., sore throat) with objective symptoms (e.g., temperature, congestion) and compares these DiagnosisGPT applies CoD to standardize this assessment process and compares these implicit and explicit symptoms against a disease database to develop a diagnostic reasoning process, a differential diagnosis with confidence distributions, and a final diagnosis.
Similarly, CoD enables a stepwise diagnostic approach with knowledge retrieval and inference based on confidence distribution over the candidate diseases.
The framework thus ensures that the diagnostic inference of CoD is grounded in a provided knowledge base, highlighting its potential flexibility in handling emerging medical scenarios.

\textbf{Agent Design:}
From the high level, CoD takes a structured Symptom Abstraction synthesized from the patient’s explicit symptoms as its inputs, and outputs a diagnostic inference with an optional confidence score.
The workflow of CoD contains three stages:

\begin{itemize}[leftmargin=*]
\setlength\itemsep{0em}

\item[] \textbf{1) Target Disease Retrieval:} In this stage, a list of candidate diseases that are most relevant to the input Symptom Abstraction is retrieved. 
These candidate diseases are ranked based on their similarity with the Symptom Abstraction in the embedding space.
This step significantly narrows down the diagnostic scope, focusing computational resources on the most likely conditions.
\item[] \textbf{2) Diagnostic Reasoning:} Once identified, the candidate diseases are assessed against the input symptoms.
Specifically, a prompt including the input symptom, the candidate diseases, and the description of each candidate disease is used to query the backbone LLM for a confidence score estimated for each candidate disease.
The descriptions for the candidate diseases are stored in an internal knowledge base and can be easily retrieved based on the disease name.
In the original CoD paper, the backbone LLM in this stage is fine-tuned on a specially created dataset to learn the mapping from the input symptoms and the candidate diseases to the confidence scores.
In our experiments, when evaluating CoD with foundation LLMs (without the special fine-tuning of CoD), we skip this step by directly inferring the most likely disease without generating confidence scores.
\item[] \textbf{3) Confidence Assessment:}
In this final stage, CoD establishes a disease confidence distribution that reflects the model's diagnostic confidence across candidates. 
This distribution not only guides decision-making but also provides an interpretation of the inference results.
Again, we simplify the pipeline of CoD in our experiments by directly producing the inference results on the candidate diseases and the input symptoms.

\end{itemize}

\textbf{Key Features and Potential Gain from o1 Integration:}
The key features of the CoD framework are its integration of confidence distribution and knowledge retrieval.
These two mechanisms critically enhance the internal reasoning of the agent and the accuracy of the diagnostic process, respectively.
The confidence distribution assigns probabilities to potential diagnoses, providing clinicians with clear insights into the model’s confidence level for each diagnostic option.
Knowledge retrieval allows the model to access a comprehensive database of medical information to support each diagnostic stage.
The integration of LLMs with strong reasoning capabilities, such as the o1 model, may further enhance the CoD's decision-making for knowledge retrieval and diagnostic reasoning.
In practice, the o1 model can potentially analyze more complex medical scenarios by considering additional factors such as patient history, co-morbidities, and even potential treatment complications.
This expanded analytical scope, enabled by the incorporation of o1, allows CoD to make nuanced adjustments to the confidence distribution, dynamically weighting factors based on their clinical relevance and urgency.

\begin{figure}[H]
\centering
\includegraphics[width=0.6\linewidth]{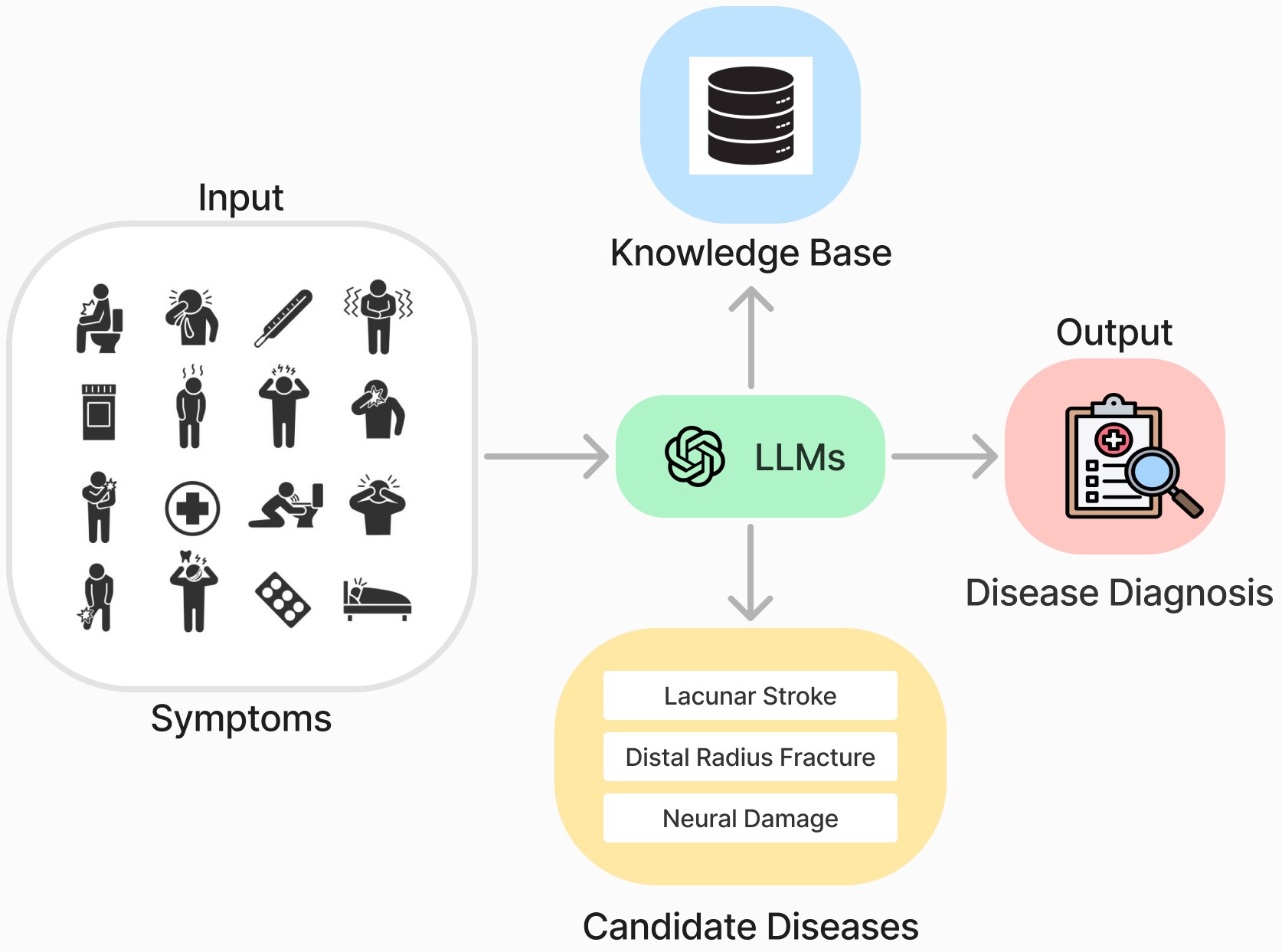}
\label{codagent_pipe}
\caption{Overview of our modified CoDAgent: Based on the input Symptoms and the Candidate Diseases, the LLM backbone model will make disease diagnosis.}
\end{figure}

\subsubsection{Experiment Design}
\textbf{Agent Settings} In this experiment, we adjust the diagnostic framework of CoD to make it more suitable for foundation LLMs.
In Step 2, we deviate from the original method of CoD that utilizes embedding similarity to select the top-k relevant diseases.
Instead, we retrieve all diseases in the candidate pool provided in the dataset and the relevant descriptions and clinical summaries for each disease. These descriptions are then inputted into the LLM, which performs a ranking operation across all candidate diseases. By evaluating each disease's relevance to the extracted symptoms, the model ranks the candidates and identifies the top-ranked disease. This ranked selection substitutes for the confidence level estimation traditionally used, providing a more interpretable and model-driven method for candidate prioritization. 

\textbf{Datasets and Metrics}
The datasets in use here are Dxbench, Dxy, and Muzhi. Muzhi and Dxy are built based on real doctor-patient consultation. DxBench is built based on the MedDialog dataset, which contains real doctor-patient dialogues.

\begin{itemize}[leftmargin=*]
\setlength\itemsep{0em}
\item[]  \textbf{1) Dxy Dataset} evaluated diagnostic models in medical AI, using real doctor-patient dialogue data. It focuses on assessing models’ ability to predict diseases based on symptoms, allowing for both multiple-choice answers and targeted symptom inquiries. For our evaluation of the Dxy dataset, we conduct tests on a subset of 120 cases. The results are divided into three folds. For each group, we calculate the mean diagnostic accuracy. The overall performance metric is derived by averaging these three group means and calculating the standard deviation across them, providing a robust measure of model performance variability across different subsets.
\item[] \textbf{2) DxBench Dataset} was designed to rigorously test diagnostic models in medical AI, using real doctor-patient dialogues covering a wide range of 461 diseases across 15 medical specialties. It challenges models to accurately diagnose based on symptoms, enabling both multiple-choice responses and symptom inquiries to improve prediction accuracy. Again we sample 120 sample cases and compute standard deviations like we did in Dxy Dataset. 
\item[] \textbf{3) Muzhi Dataset} was built from real doctor-patient consultations and was used to test the ability of models to predict diseases based on an initial set of symptoms, with the option for follow-up inquiries. Muzhi primarily employs multiple-choice questions to guide models in selecting the most probable diagnosis or in requesting additional symptom information to improve diagnostic accuracy. For our evaluation of the Muzhi Dataset, we conduct the same as previously.
\end{itemize}

\textbf{Model Choices} We test three models: GPT-4, and o1.
In step 2, the model is replaced by the OpenAI backbone models, including both GPT-4 and o1-preview, using input data consisting of patient symptoms and a list of candidate diseases. The models process this data in a multiple-choice question format to select the most likely diagnosis.

\subsubsection{Experiment Result}

\begin{table}[h!]
\centering
\caption{Accuracy of CoDAgent with GPT-4 and o1 as the backbone LLM on Dxy, DxBench, and Muzhi datasets
with the average runtime per entry. The backbone `o1' here denotes that the backbone LLM of all four agents
involved has been replaced with o1.}
% \zhen{No need to say o1-all since we don't have other o1 variants.}}
\begin{tabularx}{\textwidth} {l X X X X X X}
\toprule
& \multicolumn{2}{c}{\textbf{Dxy}} & \multicolumn{2}{c}{\textbf{DxyBench}} & \multicolumn{2}{c}{\textbf{Muzhi}} \\
\cmidrule(lr){2-3} \cmidrule(lr){4-5} \cmidrule(lr){6-7}
\textbf{Backbone} & \textbf{Accuracy(\%)} & \textbf{Runtime(s)} & \textbf{Accuracy(\%)} & \textbf{Runtime(s)}& \textbf{Accuracy(\%)} & \textbf{Runtime(s)} \\
\midrule
GPT-4 &53.04 $\pm$ 10.05    & 0.36  &  55.00$\pm$6.12 & 0.36 &  59.17   $\pm$   5.14 & 0.39        \\ 
o1  & 63.22   $\pm$ 7.88    & 3.54  & 57.53   $\pm$    9.35 & 4.44 &  60.00   $\pm$  3.54 &  4.42           \\
\bottomrule
\end{tabularx}
\label{tab1:agentcod}
\end{table}

\textbf{O1 outperforms GPT-4 on all datasets.} For the dataset Dxy, o1 reached 63.22\% (±7.88) accuracy which outperfrom GPT-4 has 53.04\% (±10.05). For the dataset DxyBench, o1 reached 57.53\% (±9.35) which also outperformed the GPT-4 has 55.00\%(±6.12). For the dataset Muzhi, o1 has 60.00\%(±3.54) also outperformed the GPT-4 has 59.17 \%(±5.14). These comparisons showed that o1 has better reasoning ability in the real doctor-patient datasets.
This margin highlights the advanced reasoning and analytical capabilities of the o1 model, which appears to better capture the nuances of real-world doctor-patient dialogue in diagnosis.

\textbf{O1 requires more runtime compared to the GPT-4.} The experimental results indicate that, while o1 consistently outperforms GPT-4 in terms of diagnostic accuracy across all datasets, this improvement comes at the cost of increased runtime. Specifically, o1’s advanced reasoning and multi-step analytical capabilities require substantially more computational resources, resulting in longer processing times. For instance, in the Dxy dataset, o1’s runtime per entry is notably higher than GPT-4’s, which may impact its practicality in real-time clinical scenarios where rapid decision-making is essential. This increased runtime is a trade-off inherent in o1’s complex reasoning framework. While o1’s enhanced accuracy makes it highly suitable for applications that prioritize diagnostic precision and depth of reasoning, the additional computational overhead may restrict its feasibility in time-sensitive clinical environments.

\textbf{O1 is more stable than GPT-4.} The o1 backbone also exhibited a smaller standard deviation (std), indicating that the model generally performs more consistently on the Dxy and DxyBench datasets. In the medical field, stable and consistent diagnostic performance is highly valuable, as it brings the model’s behavior closer to that of human doctors by providing more reliable diagnoses across varied cases.

\subsection{MedAgents}
\textbf{Overview:} MedAgents is a medical collaborative agent based on role-play.
It simulates the real clinical environment where a group of experts with diverse expertise discuss a patient's description of symptoms and reach a consensus on diagnosis and treatments.
The pipeline of MedAgents is composed of five major stages: gathering a group of experts, symptom analysis by each expert, summarization of individual expert's analysis, consultation for mutual agreements, and final decision-making \cite{tang2023medagents}.
The simulation of expert discussions with zero-shot reasoning of MedAgents enables it to outperform simple single-agent design on medical datasets and effectively harness latent medical knowledge.

\textbf{Agent Design:} From a high level, MedAgents takes clinical questions and patient data as structured inputs, synthesizing insights from various domain experts. 
The output is a consensus diagnostic inference after possibly multiple rounds of ``discussion'' between the experts.
The MedAgents workflow consists of five stages:

\begin{itemize}[leftmargin=*]
\setlength\itemsep{0em}

\item[] \textbf{1) Expert Gathering:}   MedAgents initiates by selecting virtual experts from various medical domains relevant to the clinical question. 
This is achieved by querying the backbone LLM with the input description of the patient's symptoms and an instruction to create five experts based on the description.
Note that the experts selected by the backbone LLM can be arbitrary without any pool of candidates.
\item[] \textbf{2) Analysis Proposition:} In this step, each domain expert independently provides an analysis of the clinical question, examining symptoms and potential diagnoses from their unique perspective, contributing distinguished insights for accurate reasoning.
Specifically, for each expert, the backbone LLM is prompted with the role of the expert and a description such as the specialized knowledge area of this expert.               
\item[] \textbf{3) Report Summarization:}  
This stage aims to refine the preliminary analysis and generate a tentative summary for voting.
The backbone LLM is prompted with the analysis generated by the five selected experts and an instruction to aggregate these analyses.     
\item[] \textbf{4) Collaborative Consultation:} 
This step aims to reach a consensus between the experts and to produce a finalized report.
The experts are engaged in a ``discussion'' around the summary report.
If a consensus (via voting) can not reached, the minority expert will be asked to refine the summary report based on his/her expertise.
The refined summary report will be voted on by all the experts again.
This voting-refinement procedure is iterated until a consensus or the maximum allowed iteration is reached (whichever comes first).
\item[] \textbf{5) Decision Making:}      A final consensus-based diagnosis is generated by prompting the backbone LLM with the patient's description of the symptoms and the summary report.
Both the finalized clinical decision (a symbolic option choice for multi-choice questions) and the summary report supporting the decision will be provided to the user.

\end{itemize}

\textbf{Key Features and Potential Gain from o1 Integration:} MedAgents is a role-play-based multi-agent system mirroring real-world clinical workflows.
It leverages a symptom-based agent-selection mechanism to identify relevant, field-specific agents and facilitate an in-depth collaborative consultation process. The integration of the o1 model is anticipated to enhance this collaborative diagnostic process by contributing advanced reasoning capabilities.
With o1’s capacity to incorporate complex factors, such as patient history and co-morbidities, MedAgents aims to achieve increased diagnostic precision and adaptability, yielding more accurate disease diagnosis.

\begin{figure}[h!]
\centering
\includegraphics[width=\linewidth]{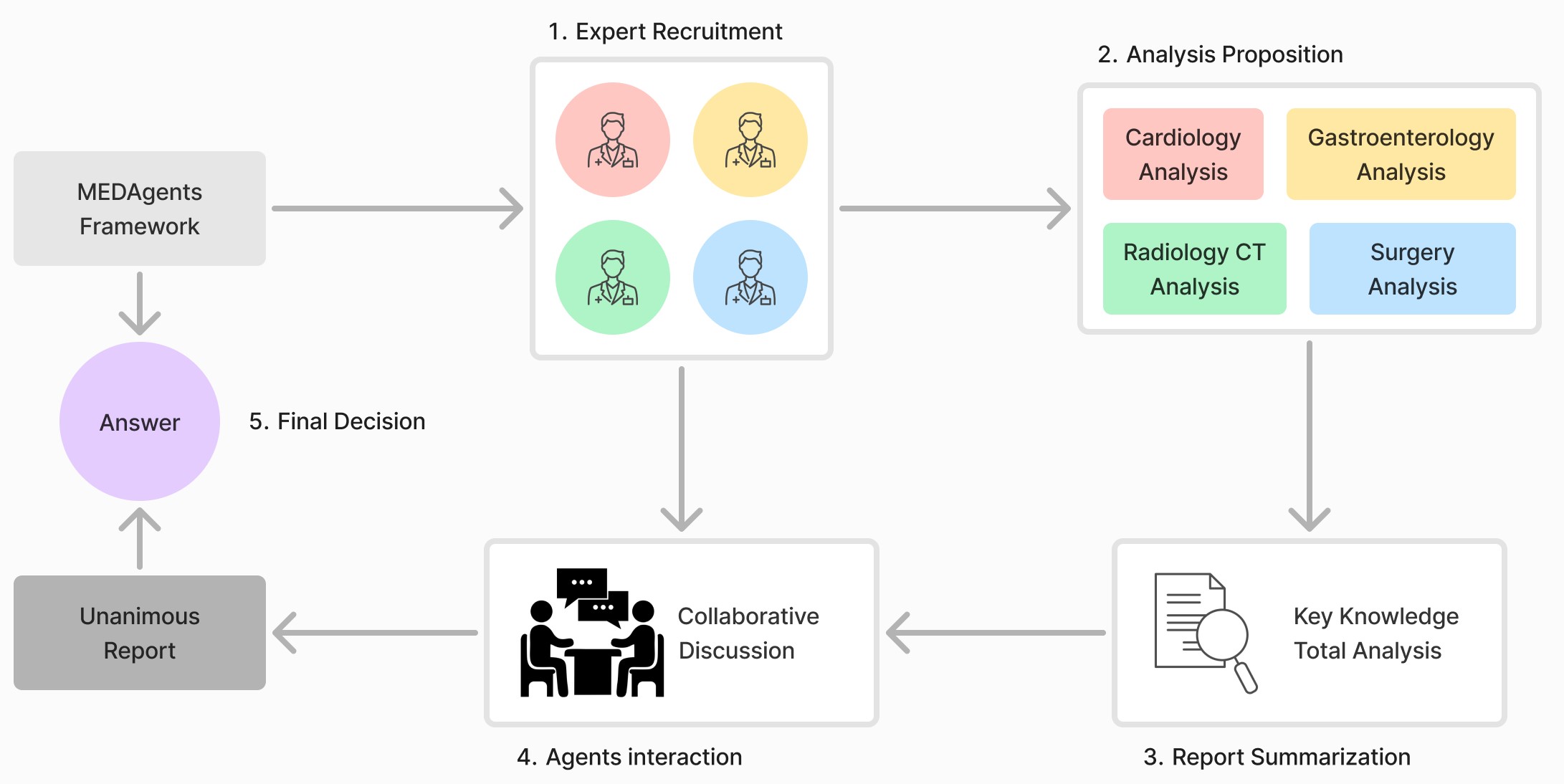}
\label{medagent_pipe}
\caption{Overview of the MedAgents pipeline with five stages: 1) expert recruitment that assigns a group of different experts related to the clinical question, 2) analysis proposition where each expert creates an analysis according to the clinical question based on its assigned role, 3) report summarization where the analysis by all recruited experts are aggregated into one, 4) collaborative consultation that facilitates expert review and iterative modifications until consensus is reached on a final report, 5) final decision where a clinical decision is made based on the summary report.
}
\end{figure}

\subsubsection{Experiment Design}
\textbf{Agent Setting}
In each step, we replace the MedAgents backbone model to study the influence of the backbone model on task performance. By flexibly utilizing various backbones during expert gathering, analysis, report summarization, collaborative consultation, and decision-making, MedAgents leverages the unique strengths of each model, enhancing diagnostic accuracy and adaptability.

\textbf{Datasets and Metrics}
MedQA and MedMCQA are high-caliber datasets curated to rigorously benchmark the diagnostic reasoning and specialized medical knowledge processing abilities of large language models (LLMs) within clinical and academic healthcare contexts.

\begin{itemize}[leftmargin=*]
\setlength\itemsep{0em}
\item[] \textbf{1) MedQA:} The MedQA dataset, sourced from the U.S. Medical Licensing Examination (USMLE), comprises 215 multiple-choice questions (MCQs), each with four answer options. To optimize efficiency and computational feasibility, we randomly selected 120 questions from this dataset. This selection ensures a broad representation of medical topics while minimizing processing requirements.

\item[] \textbf{2) MedMCQA:} Derived from various medical examinations like NEET-PG and AIIMS-PG, the MedMCQA dataset comprises over 194,000 multiple-choice questions (MCQs), each with four answer options. To optimize efficiency and computational feasibility, we randomly selected a subset of questions from this extensive dataset. This selection ensures a broad representation of medical topics while keeping processing requirements manageable.

\end{itemize}

\textbf{Model Choices}
In this study, we compare GPT-4 and the o1 model (specifically, o1-preview) as the core engines powering MedAgents language agents. GPT-4 is used as the baseline due to its established strengths in language comprehension and reasoning. The o1 model, recognized for its advanced multi-step reasoning and adaptability, is explored as a possible upgrade aimed at enhancing diagnostic accuracy and reliability. By substituting GPT-4 with o1, we assess how effectively the o1 model can boost performance in complex, high-stakes medical scenarios, with a particular emphasis on deepened reasoning and reduced diagnostic variability.

\subsubsection{Experiment Result}

\begin{table}[H]
\centering
\caption{Accuracy of MeDAgent with GPT-4 and o1 as the backbone LLM on MedQA MedMCQA datasets
with the average runtime per entry. The backbone `o1’ here denotes that the backbone LLM of all four agents
involved has been replaced with o1. The backbone 'GPT4-EG' here denotes the backbone of change GPT-4 for the Expert Gathering step.
}
\begin{tabularx}{\textwidth} {l X X X X}
\toprule
& \multicolumn{2}{c}{\textbf{MedQA}} & \multicolumn{2}{c}{\textbf{MedMCQA}}\\
\cmidrule(lr){2-3} \cmidrule(lr){4-5}
\textbf{Backbone} & \textbf{Accuracy(\%)} & \textbf{Runtime(s)} & \textbf{Accuracy(\%)} & \textbf{Runtime(s)} \\
\midrule
GPT-4  &78.91   $\pm$ 6.92 & 64.21 &64.17  $\pm$ 6.24 & 69.42 \\ 
GPT4-EG &92.39 $\pm$ 4.21 & 221.33 & 82.50 $\pm$ 4.08 & 200.33 \\
o1  &87.50 $\pm$ 4.08 & 250.65 & 79.17 $\pm$ 1.18 & 240.21 \\
\bottomrule
\end{tabularx}
\label{tab2:medagent}
\end{table}
%MedMc数据名字typo修改. v.s.
\textbf{O1 outperforms GPT-4 in all datasets and shows more stable results} The results in Table \ref{tab2:medagent} show that o1 outperforms GPT-4 across all datasets with a clear margin, especially on the MedMcQA dataset.
% For the MedQA dataset, the o1 model achieved an accuracy of 87.50\% (±4.08), surpassing GPT-4’s accuracy of 78.91\% (±6.92). This suggests that o1 is more accurate than GPT-4 in handling the MedQA dataset.
Additionally, the smaller standard deviation (4.08 vs. 6.92 on MedQA and 1.18 vs. 6.24 on MedMCQA) for o1 indicates more consistent performance across cases, which is advantageous in medical applications where reliability is essential.
% For the MedMCQA dataset, o1 also demonstrated better performance, with an accuracy of 79.17\%, compared to GPT-4’s 64.17\%, and a smaller standard deviation (1.18 vs. 6.24).

\textbf{Change GPT-4 on expert gathering step shows better results than use o1 for all} In the GPT4-EG, 
we used GPT-4 as the backbone for the first step, Expert Gathering. 
Recall that in this step, the backbone LLM is queried with a specific prompt based on the question to select five experts.
% We observed an unexpected increase in accuracy with this change. In this initial step, we design a specific prompt based on the question-and-answer options to query the model and select five experts. 
As shown in Table \ref{tab2:medagent}, GPT-4 performed better in expert selection than o1, identifying more relevant experts to support the following steps.
Our conjecture is that the expert selection is a relatively simple task where o1 may perform redundant reasoning than needed, which complicates the problem.
This will be further investigated in our future work.

% To further investigate this, we wanted to examine the differences between GPT-4’s selections and those made by o1, as expert field selection is the only variable between them. After reviewing several samples, we observed that GPT-4 tends to select expert fields more closely related to the question and options compared to o1. 
% \zhen{Need to give more details.}

\subsection{AgentClinic}
\textbf{Overview:} AgentClinic \cite{schmidgall2024agentclinicmultimodalagentbenchmark} is a multi-agent system simulating real clinical environments.
Different from MEDAgents which uses a role-play-based approach, AgentClinic involves multiple individual agents interacting directly with each other.
The AgentClinic system includes four agents: a patient agent, a doctor agent, a measurement agent, and a moderator agent.
The patient agent provides symptoms like a real human patient, while the doctor agent asks questions and consults the measurement agent for test results to refine its diagnosis. The moderator agent then validates the doctor’s conclusion, simulating the collaborative, multi-step process typical of real healthcare scenarios.
Such a simulation of the real-world clinical environments aims to evaluate how well AI-driven agents manage patient interactions, respond to evolving cases, and make decisions that mirror the flow of actual clinical practice. In a simulated scenario conducted purely through text-based interactions, the patient agent presents a set of symptoms without access to the final diagnosis, requiring the doctor agent to actively engage by asking diagnostic questions, conducting virtual tests, and working toward an accurate diagnosis.

\textbf{Agent Design:} AgentClinic is a multi-agent diagnostic system designed to emulate real-life clinical workflows through layered information processing. Following a modular design, it comprises four core agents—the patient, doctor, measurement, and moderator agents—each assigned specific roles to build a comprehensive diagnosis.
The patient agent initiates the process by providing symptom data.
The doctor agent interprets and investigates through questions and virtual tests.
For objective data, the doctor agent requests test results from the measurement agent.
Finally, the moderator agent verifies the diagnosis, ensuring that each step, from data collection to diagnosis validation, mirrors the structured flow of real-world clinical consultations.

\begin{itemize}[leftmargin=*]
\setlength\itemsep{0em}
    
\item[] \textbf{1) Patient Interaction and Information Gathering:} The process begins with the patient agent, simulating real interactions by sharing symptoms, and medical history, and responding to the doctor’s questions. This is a purely text-based conversational process (See \ref{app:agentclinic}).
The patient agent doesn’t know the end diagnosis, meaning the doctor agent has to dig deep, asking follow-up questions to uncover critical information. This realistic setup requires the doctor agent to fully engage, just like a physician would, creating a lifelike environment\cite{ely2000}.

\item[] \textbf{2) Diagnostic Reasoning and Hypothesis Generation:} Once the initial information is gathered, the doctor agent begins the reasoning phase under a system prompt that defines its diagnostic role. This prompt directs the agent to analyze patient data step-by-step, synthesizing symptoms, test results, and medical history to produce a ranked list of possible diagnoses. The input consists of patient data gathered from the conversation with the patient agent, including responses and test results from the measurement agent, while the output is a prioritized list of diagnoses refined iteratively as new data is considered. The conversation concludes either when a diagnosis is reached or when the maximum iterations are completed, at which point the doctor agent provides the most probable diagnosis based on prior interactions.

\item[] \textbf{3) Objective Testing and Measurement Validation:} For a more thorough evaluation, the doctor agent can turn to the measurement agent for objective data—such as lab results or imaging studies. The addition of the measurement agent is critical to replicating the multi-dimensional approach required in actual diagnostics.
This step is essential for balancing subjective patient-reported symptoms with hard clinical evidence, giving the doctor agent a more accurate basis for its conclusions. 

\item[] \textbf{4) Diagnosis Validation and Moderator Review:} Finally, the moderator agent steps in to check the doctor’s work. It compares the diagnosis against the known outcome, ensuring accuracy even if the doctor’s phrasing is slightly off. This layer of oversight completes the workflow by confirming that the AI’s reasoning aligns with clinical standards and expectations. If the workflow fails the examination, the system will output "diagnosis is incorrect" to avoid misinforming clinicians and ensure any discrepancies are clearly flagged for further review.

\end{itemize}

\textbf{Key Features and Potential of o1 Model Integration:} 
AgentClinic’s defining qualities include its lifelike patient interaction, structured decision-making flow, and ability to assess AI performance in biased scenarios through reasoning during multi-hop QA.
The integration of the o1 model could take this setup even further. By enhancing the doctor agent’s ability to reason through complex cases and make more targeted test selections, o1 could help the AI reach more accurate conclusions. This advanced layer of reasoning may enable the doctor agent to surpass human-level performance in diagnostic accuracy.

\begin{figure}[h!]
\centering
\includegraphics[width=\linewidth]{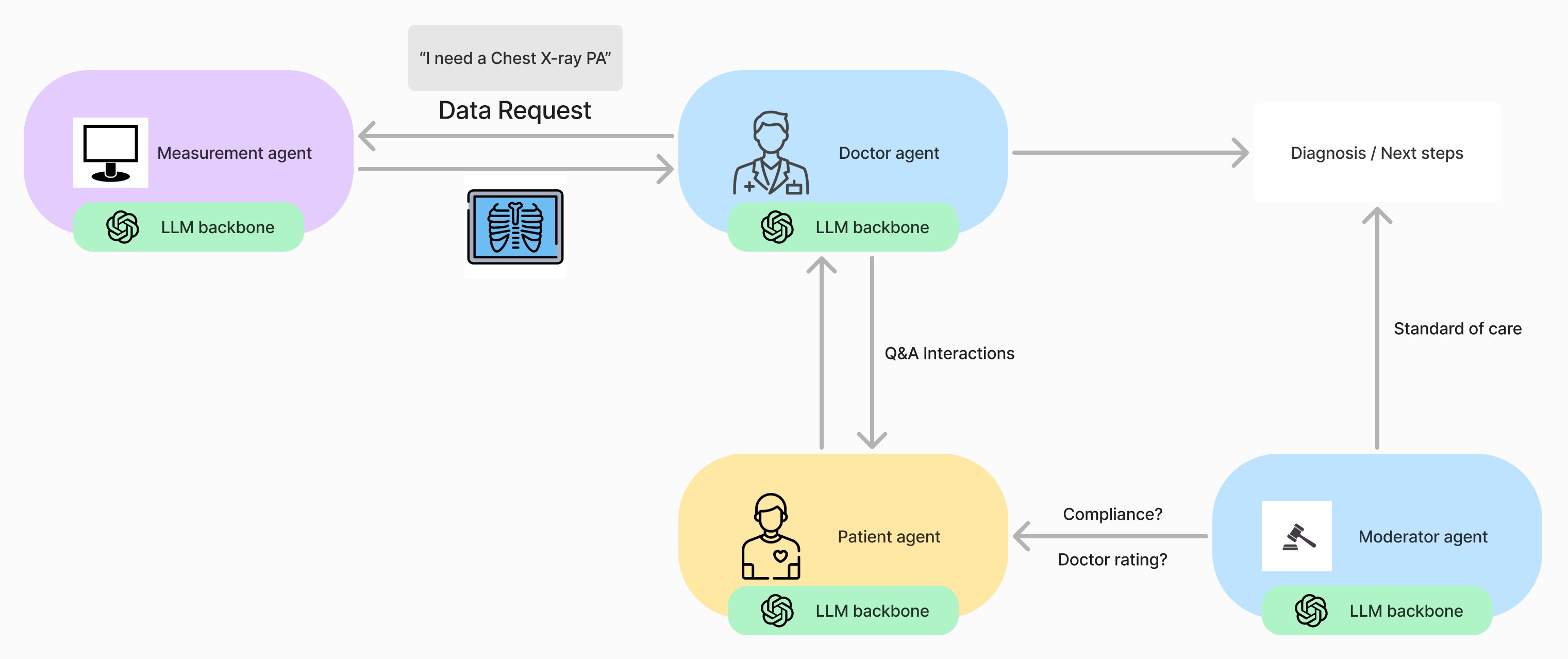}
\caption{Overview of the AgentClinic pipeline, the interaction between four core agents—Measurement, Doctor, Patient, and Moderator—each supported by an LLM backbone. The doctor agent engages with the patient agent to gather symptoms and requests additional data (e.g., X-rays) from the measurement agent to aid in diagnosis. Once a diagnosis is reached, the moderator agent verifies it against standard care practices, ensuring clinical accuracy and compliance.}
\label{agentclinic_pipe}
\end{figure}

\subsubsection{Experiment Design}
\textbf{Agent Settings:} In our study, we evaluate and compare the performance of two different LLM backbones, GPT-4 and o1-preview, as the core models for language agents within the AgentClinic framework. This setup involves four key agents—doctor, patient, measurement, and moderator—each responsible for specific tasks in the diagnostic process. We perform two sets of experiments: one where all agents are powered by GPT-4 and another with o1-preview as the backbone. By consistently replacing the backbone model across all agents, we aim to observe how each model impacts overall diagnostic accuracy, reasoning consistency, and adaptability within simulated medical decision-making tasks.

\textbf{Datasets and Metrics:} We utilize two primary datasets to benchmark the diagnostic reasoning and specialized knowledge capabilities of each model in clinical contexts: MedQA and the NEJM case challenge dataset.
\begin{itemize}[leftmargin=*]
\setlength\itemsep{0em}
\item \textbf{1) MedQA:} Derived from the U.S. Medical Licensing Examination (USMLE), the MedQA dataset contains 215 multiple-choice questions (MCQs) with four answer options. For efficiency and computational feasibility, we randomly sampled 120 entries from this dataset, ensuring a broad representation of medical topics while minimizing processing requirements.
\item \textbf{2) NEJM:} This dataset consists of 120 clinical cases that require complex diagnostic reasoning, often using multimodal data (text and images). Given that o1-preview does not support image processing, we conducted all experiments without the images to ensure comparability between models. Although this limitation may impact overall accuracy, the NEJM dataset still provides valuable insights into each model’s textual reasoning and diagnostic capabilities.
\end{itemize}

\textbf{Model Choices:} In this study, we compare GPT-4 and o1 (more specifically o1-preview) as the backbone models for AgentClinic’s language agents.
GPT-4 serves as the baseline model, given its established capabilities in language understanding and reasoning. The o1 model, known for its enhanced multi-step reasoning and adaptability, is tested as a potential upgrade to improve diagnostic accuracy and reliability. By replacing GPT-4 with o1, we investigate the extent to which the o1 model can enhance performance in complex, high-stakes medical scenarios, particularly focusing on improved reasoning depth and reduced diagnostic variability.

\subsubsection{Experiment Result}

\begin{table}[h!]
\centering
\caption{Performance accuracy of AgentClinic with GPT-4 and o1-preview as the backbone LLMs on the MedQA and NEJM datasets, along with the average runtime per entry. In this setup, 'o1-all' indicates that all four agents utilize o1 as the backbone LLM. 'o1-doctor' denotes that only the doctor agent is powered by o1, with the remaining agents using GPT-4, while 'o1-patient' specifies that only the patient agent utilizes o1 as the backbone LLM, with the others using GPT-4.}
\begin{tabularx}{\textwidth} {l X X X X}
\toprule
& \multicolumn{2}{c}{\textbf{MedQA}} & \multicolumn{2}{c}{\textbf{NEJM}} \\
\cmidrule(lr){2-3} \cmidrule(lr){4-5}
\textbf{Backbone} & \textbf{ Accuracy (\%) } & \textbf{Runtime (s)} & \textbf{ Accuracy (\%)} & \textbf{Runtime (s)} \\
\midrule
GPT-4 & 59.17 ± 3.12 & 89.93 & 25.00 ± 2.04 & 89.91\\
o1-doctor & 77.50 ± 5.00 & 112.57 & 38.33 ± 5.77 & 138.06\\
o1-patient & 58.33 ± 8.78 & 207.57 & 23.33 ± 1.44 & 202.96\\
o1-all & 75.83 ± 1.18 & 205.76 & 35.83 ± 6.24 & 197.94\\
\bottomrule
\end{tabularx}
\label{tab3:agentclinic}
\end{table}

\textbf{Selective replacement of the doctor agent yields significant improvements in diagnostic accuracy.} The experimental results show that o1-based backbones, particularly o1-all and o1-doctor, lead to substantial improvements in diagnostic accuracy for AgentClinic compared to GPT-4. On the MedQA dataset, o1-all achieved an accuracy of 75.83\% (±1.18), a considerable increase from GPT-4's 59.17\% (±3.12). For NEJM, o1-all reached 35.83\% (±6.24) versus GPT-4’s 25.00\% (±2.04), demonstrating enhanced capabilities in handling both structured and complex medical tasks. Notably, o1-doctor achieved the highest accuracy on both datasets, with 77.50\% on MedQA and 38.33\% on NEJM, suggesting that upgrading only the doctor agent can capture the diagnostic advantages of o1’s advanced reasoning without additional computational complexity in other agents. The observed higher variance in o1's performance compared to GPT-4 can be attributed to the characteristics of our sampling and testing methodology. Specifically, the variance naturally fluctuates in proportion to the overall accuracy of the testing set, reflecting the inherent variability in the sampled cases.

\textbf{Strategically replacing only the doctor agent provides a balance between accuracy and computational efficiency.} While o1-all demonstrated higher accuracy, it also required significantly more computational resources, with runtime per entry more than twice that of GPT-4 for both datasets. In contrast, o1-doctor showed a smaller runtime increase (112.57s for MedQA and 138.06s for NEJM), offering an efficient middle ground with high diagnostic performance but a more manageable computational demand. This improved balance is likely due to the doctor agent’s central role in diagnostic reasoning, making it the most effective place to incorporate o1’s reasoning capabilities. By contrast, o1-patient—where only the patient agent uses o1—showed minimal accuracy improvements and similar runtimes to o1-all, indicating that enhancing only the patient agent does not substantially affect diagnostic outcomes due to its more passive, information-providing role.

\section{Conclusion}
This study investigated the integration of the o1 model into various LLM-based agents within the medical domain, replacing traditional backbone models like GPT-4. Our experiments demonstrated that the o1 model significantly enhances performance across multiple medical reasoning tasks. In the CoD Agent framework, o1-preview not only matched but often exceeded GPT-4 in accuracy, showing more consistent diagnostic results with lower standard deviations. In the MedAgent framework, o1-preview performs better on the MedQA and MedMCQA datasets than GPT-4 and demonstrates greater stability. Similarly, in the AgentClinic simulations, o1-preview achieved higher mean accuracies on both the MedQA and NEJM datasets compared to GPT-4, indicating superior diagnostic capabilities even in complex, multimodal scenarios.

The future of healthcare envisions a sophisticated, multi-agent LLM framework that enables seamless collaboration across departments and specialties. Leveraging the o1 model's advanced Chain-of-Thought (CoT) reasoning and Retrieval-Augmented Generation (RAG) techniques, LLM agents can enhance diagnostic workflows and cross-disciplinary consultations. By embedding CoT reasoning directly into its training process and incorporating reinforcement learning from human feedback, o1-preview demonstrates enhanced multi-step reasoning abilities, effectively managing complex diagnostic queries, simulating clinical workflows, and adapting to dynamic environments like ICUs. Through specialized agents for diagnosis, monitoring, and treatment coordination, future hospital systems can dynamically adapt to real-time patient data, enhance diagnostic precision, and optimize treatment pathways. This collaborative, multi-agent approach promises to transform medical decision-making into a comprehensive, efficient process, aligning interdisciplinary insights to ultimately improve patient outcomes.

However, the experiments also highlighted some limitations. The current version of o1-preview lacks support for multimodal data processing, affecting its performance on datasets that include imaging data, such as the NEJM dataset. In the medical field, the ability to process multimodal inputs is essential, as complex clinical tasks often require the integration of various data types, such as radiology images/videos, clinical notes, and speech data, to provide accurate and comprehensive insights. Leveraging the advanced reasoning capabilities of the o1-preview model as the central component of a multimodal agent framework could effectively address the complexities involved in interpreting diverse clinical factors across various scenarios. Additionally, the increased computational runtime with o1-preview indicates a trade-off between accuracy and efficiency, which could be a concern in time-sensitive clinical settings.

Future research should focus on addressing these limitations by integrating the o1 model into a multimodal multi-agent framework. This integration would enhance diagnostic capabilities by positioning the o1 model as a task planner that utilizes advanced reasoning to coordinate and assign subtasks across different modalities for various expert models. Furthermore, optimizing computational efficiency without compromising performance is also crucial for practical applications in real-world clinical environments.

In conclusion, replacing traditional backbone models with the o1 model in medical AI agents offers significant practical benefits. The o1 model's superior reasoning, adaptability, and consistent performance across diverse clinical tasks position it as a promising tool for advancing healthcare delivery. By continuing to refine models like o1 and integrating them into clinical practice, we move closer to developing AI agents that can meet the complex demands of modern medicine, ultimately enhancing patient care and outcomes.

%Bibliography
\bibliographystyle{unsrt}  
\bibliography{references}  

\begin{thebibliography}{10}

\bibitem{zhong2024evaluation}
Tianyang Zhong, Zhengliang Liu, Yi~Pan, Yutong Zhang, Yifan Zhou, Shizhe Liang, Zihao Wu, Yanjun Lyu, Peng Shu, Xiaowei Yu, et~al.
\newblock Evaluation of openai o1: Opportunities and challenges of agi.
\newblock {\em arXiv preprint arXiv:2409.18486}, 2024.

\bibitem{liu2023holistic}
Zhengliang Liu, Hanqi Jiang, Tianyang Zhong, Zihao Wu, Chong Ma, Yiwei Li, Xiaowei Yu, Yutong Zhang, Yi~Pan, Peng Shu, et~al.
\newblock Holistic evaluation of gpt-4v for biomedical imaging.
\newblock {\em arXiv preprint arXiv:2312.05256}, 2023.

\bibitem{nori2023capabilitiesgpt4medicalchallenge}
Harsha Nori, Nicholas King, Scott~Mayer McKinney, Dean Carignan, and Eric Horvitz.
\newblock Capabilities of gpt-4 on medical challenge problems, 2023.

\bibitem{holmes2023evaluating}
Jason Holmes, Rui Peng, Yiwei Li, Jinyu Hu, Zhengliang Liu, Zihao Wu, Huan Zhao, Xi~Jiang, Wei Liu, Hong Wei, et~al.
\newblock Evaluating multiple large language models in pediatric ophthalmology.
\newblock {\em arXiv preprint arXiv:2311.04368}, 2023.

\bibitem{li2024echopulse}
Yiwei Li, Sekeun Kim, Zihao Wu, Hanqi Jiang, Yi~Pan, Pengfei Jin, Sifan Song, Yucheng Shi, Tianze Yang, Tianming Liu, et~al.
\newblock Echopulse: Ecg controlled echocardio-grams video generation.
\newblock {\em arXiv preprint arXiv:2410.03143}, 2024.

\bibitem{nori2023generalistfoundationmodelsoutcompete}
Harsha Nori, Yin~Tat Lee, Sheng Zhang, Dean Carignan, Richard Edgar, Nicolo Fusi, Nicholas King, Jonathan Larson, Yuanzhi Li, Weishung Liu, Renqian Luo, Scott~Mayer McKinney, Robert~Osazuwa Ness, Hoifung Poon, Tao Qin, Naoto Usuyama, Chris White, and Eric Horvitz.
\newblock Can generalist foundation models outcompete special-purpose tuning? case study in medicine, 2023.

\bibitem{saab2024capabilitiesgeminimodelsmedicine}
Khaled Saab, Tao Tu, Wei-Hung Weng, Ryutaro Tanno, David Stutz, Ellery Wulczyn, Fan Zhang, Tim Strother, Chunjong Park, Elahe Vedadi, Juanma~Zambrano Chaves, Szu-Yeu Hu, Mike Schaekermann, Aishwarya Kamath, Yong Cheng, David G.~T. Barrett, Cathy Cheung, Basil Mustafa, Anil Palepu, Daniel McDuff, Le~Hou, Tomer Golany, Luyang Liu, Jean baptiste Alayrac, Neil Houlsby, Nenad Tomasev, Jan Freyberg, Charles Lau, Jonas Kemp, Jeremy Lai, Shekoofeh Azizi, Kimberly Kanada, SiWai Man, Kavita Kulkarni, Ruoxi Sun, Siamak Shakeri, Luheng He, Ben Caine, Albert Webson, Natasha Latysheva, Melvin Johnson, Philip Mansfield, Jian Lu, Ehud Rivlin, Jesper Anderson, Bradley Green, Renee Wong, Jonathan Krause, Jonathon Shlens, Ewa Dominowska, S.~M.~Ali Eslami, Katherine Chou, Claire Cui, Oriol Vinyals, Koray Kavukcuoglu, James Manyika, Jeff Dean, Demis Hassabis, Yossi Matias, Dale Webster, Joelle Barral, Greg Corrado, Christopher Semturs, S.~Sara Mahdavi, Juraj Gottweis, Alan Karthikesalingam, and Vivek Natarajan.
\newblock Capabilities of gemini models in medicine, 2024.

\bibitem{li2023artificial}
Xiang Li, Lu~Zhang, Zihao Wu, Zhengliang Liu, Lin Zhao, Yixuan Yuan, Jun Liu, Gang Li, Dajiang Zhu, Pingkun Yan, et~al.
\newblock Artificial general intelligence for medical imaging.
\newblock {\em arXiv preprint arXiv:2306.05480}, 2023.

\bibitem{liévin2023largelanguagemodelsreason}
Valentin Liévin, Christoffer~Egeberg Hother, Andreas~Geert Motzfeldt, and Ole Winther.
\newblock Can large language models reason about medical questions?, 2023.

\bibitem{densen2011challenges}
Peter Densen.
\newblock Challenges and opportunities facing medical education.
\newblock {\em Transactions of the American Clinical and Climatological Association}, 122:48, 2011.

\bibitem{10.1093/jamia/ocac064}
Jens~Michael Boss, Gagan Narula, Christian Straessle, Jan Willms, Jan Azzati, Dominique Brodbeck, Rahel Luethy, Susanne Suter, Christof Buehler, Carl Muroi, David~Jule Mack, Marko Seric, Daniel Baumann, and Emanuela Keller.
\newblock {ICU Cockpit: a platform for collecting multimodal waveform data, AI-based computational disease modeling and real-time decision support in the intensive care unit}.
\newblock {\em Journal of the American Medical Informatics Association}, 29(7):1286--1291, 05 2022.

\bibitem{shi2024ehragent}
Wenqi Shi, Ran Xu, Yuchen Zhuang, Yue Yu, Jieyu Zhang, Hang Wu, Yuanda Zhu, Joyce Ho, Carl Yang, and May~D Wang.
\newblock Ehragent: Code empowers large language models for complex tabular reasoning on electronic health records.
\newblock {\em EMNLP}, 2024.

\bibitem{abbasian2024conversationalhealthagentspersonalized}
Mahyar Abbasian, Iman Azimi, Amir~M. Rahmani, and Ramesh Jain.
\newblock Conversational health agents: A personalized llm-powered agent framework, 2024.

\bibitem{tu2024conversationaldiagnosticai}
Tao Tu, Anil Palepu, Mike Schaekermann, Khaled Saab, Jan Freyberg, Ryutaro Tanno, Amy Wang, Brenna Li, Mohamed Amin, Nenad Tomasev, Shekoofeh Azizi, Karan Singhal, Yong Cheng, Le~Hou, Albert Webson, Kavita Kulkarni, S~Sara Mahdavi, Christopher Semturs, Juraj Gottweis, Joelle Barral, Katherine Chou, Greg~S Corrado, Yossi Matias, Alan Karthikesalingam, and Vivek Natarajan.
\newblock Towards conversational diagnostic ai, 2024.

\bibitem{yang2024psychogat}
Qisen Yang, Zekun Wang, Honghui Chen, Shenzhi Wang, Yifan Pu, Xin Gao, Wenhao Huang, Shiji Song, and Gao Huang.
\newblock {P}sycho{GAT}: A novel psychological measurement paradigm through interactive fiction games with {LLM} agents.
\newblock In {\em Proceedings of the 62nd Annual Meeting of the Association for Computational Linguistics (Volume 1: Long Papers)}, pages 14470--14505, Bangkok, Thailand, August 2024. Association for Computational Linguistics.

\bibitem{dai2023ad}
Haixing Dai, Yiwei Li, Zhengliang Liu, Lin Zhao, Zihao Wu, Suhang Song, Ye~Shen, Dajiang Zhu, Xiang Li, Sheng Li, et~al.
\newblock Ad-autogpt: An autonomous gpt for alzheimer's disease infodemiology.
\newblock {\em arXiv preprint arXiv:2306.10095}, 2023.

\bibitem{xiao2024comprehensivesurveylargelanguage}
Hanguang Xiao, Feizhong Zhou, Xingyue Liu, Tianqi Liu, Zhipeng Li, Xin Liu, and Xiaoxuan Huang.
\newblock A comprehensive survey of large language models and multimodal large language models in medicine, 2024.

\bibitem{parisi2022talmtoolaugmentedlanguage}
Aaron Parisi, Yao Zhao, and Noah Fiedel.
\newblock Talm: Tool augmented language models, 2022.

\bibitem{schick2023toolformer}
Timo Schick, Jane Dwivedi-Yu, Roberto Dessi, Roberta Raileanu, Maria Lomeli, Eric Hambro, Luke Zettlemoyer, Nicola Cancedda, and Thomas Scialom.
\newblock Toolformer: Language models can teach themselves to use tools.
\newblock In {\em Thirty-seventh Conference on Neural Information Processing Systems}, 2023.

\bibitem{yu2023finmemperformanceenhancedllmtrading}
Yangyang Yu, Haohang Li, Zhi Chen, Yuechen Jiang, Yang Li, Denghui Zhang, Rong Liu, Jordan~W. Suchow, and Khaldoun Khashanah.
\newblock Finmem: A performance-enhanced llm trading agent with layered memory and character design, 2023.

\bibitem{li2024agenthospitalsimulacrumhospital}
Junkai Li, Siyu Wang, Meng Zhang, Weitao Li, Yunghwei Lai, Xinhui Kang, Weizhi Ma, and Yang Liu.
\newblock Agent hospital: A simulacrum of hospital with evolvable medical agents, 2024.

\bibitem{deng2023mind2webgeneralistagentweb}
Xiang Deng, Yu~Gu, Boyuan Zheng, Shijie Chen, Samuel Stevens, Boshi Wang, Huan Sun, and Yu~Su.
\newblock Mind2web: Towards a generalist agent for the web, 2023.

\bibitem{xiang2024guardagent}
Zhen Xiang, Linzhi Zheng, Yanjie Li, Junyuan Hong, Qinbin Li, Han Xie, Jiawei Zhang, Zidi Xiong, Chulin Xie, Carl Yang, Dawn Song, and Bo~Li.
\newblock Guardagent: Safeguard llm agents by a guard agent via knowledge-enabled reasoning, 2024.

\bibitem{gur2024a}
Izzeddin Gur, Hiroki Furuta, Austin~V Huang, Mustafa Safdari, Yutaka Matsuo, Douglas Eck, and Aleksandra Faust.
\newblock A real-world webagent with planning, long context understanding, and program synthesis.
\newblock In {\em The Twelfth International Conference on Learning Representations}, 2024.

\bibitem{chen2024agentpoisonredteamingllmagents}
Zhaorun Chen, Zhen Xiang, Chaowei Xiao, Dawn Song, and Bo~Li.
\newblock Agentpoison: Red-teaming llm agents via poisoning memory or knowledge bases, 2024.

\bibitem{li2024idsagent}
Yanjie Li, Zhen Xiang, Nathaniel~D. Bastian, Dawn Song, and Bo~Li.
\newblock {IDS}-agent: An {LLM} agent for explainable intrusion detection in iot networks.
\newblock In {\em NeurIPS 2024 Workshop on Open-World Agents}, 2024.

\bibitem{wei2023chainofthoughtpromptingelicitsreasoning}
Jason Wei, Xuezhi Wang, Dale Schuurmans, Maarten Bosma, Brian Ichter, Fei Xia, Ed~Chi, Quoc Le, and Denny Zhou.
\newblock Chain-of-thought prompting elicits reasoning in large language models, 2023.

\bibitem{xie2024preliminarystudyo1medicine}
Yunfei Xie, Juncheng Wu, Haoqin Tu, Siwei Yang, Bingchen Zhao, Yongshuo Zong, Qiao Jin, Cihang Xie, and Yuyin Zhou.
\newblock A preliminary study of o1 in medicine: Are we closer to an ai doctor?, 2024.

\bibitem{chen2024chain}
Junying Chen, Chi Gui, Anningzhe Gao, Ke~Ji, Xidong Wang, Xiang Wan, and Benyou Wang.
\newblock Chain-of-diagnosis: Towards an interpretable medical agent using chain of diagnosis.
\newblock {\em arXiv preprint arXiv:2407.13301}, 2024.

\bibitem{tang2023medagents}
Xiangru Tang, Anni Zou, Zhuosheng Zhang, Ziming Li, Yilun Zhao, Xingyao Zhang, Arman Cohan, and Mark Gerstein.
\newblock Medagents: Large language models as collaborators for zero-shot medical reasoning.
\newblock {\em arXiv preprint arXiv:2311.10537}, 2023.

\bibitem{schmidgall2024agentclinicmultimodalagentbenchmark}
Samuel Schmidgall, Rojin Ziaei, Carl Harris, Eduardo Reis, Jeffrey Jopling, and Michael Moor.
\newblock Agentclinic: a multimodal agent benchmark to evaluate ai in simulated clinical environments, 2024.

\bibitem{liu2023evaluating}
Zhengliang Liu, Tianyang Zhong, Yiwei Li, Yutong Zhang, Yi~Pan, Zihao Zhao, Peixin Dong, Chao Cao, Yuxiao Liu, Peng Shu, et~al.
\newblock Evaluating large language models for radiology natural language processing.
\newblock {\em arXiv preprint arXiv:2307.13693}, 2023.

\bibitem{zhang2024generalist}
Kai Zhang, Rong Zhou, Eashan Adhikarla, Zhiling Yan, Yixin Liu, Jun Yu, Zhengliang Liu, Xun Chen, Brian~D Davison, Hui Ren, et~al.
\newblock A generalist vision--language foundation model for diverse biomedical tasks.
\newblock {\em Nature Medicine}, pages 1--13, 2024.

\bibitem{liu2023radiology}
Zhengliang Liu, Yiwei Li, Peng Shu, Aoxiao Zhong, Longtao Yang, Chao Ju, Zihao Wu, Chong Ma, Jie Luo, Cheng Chen, et~al.
\newblock Radiology-llama2: Best-in-class large language model for radiology.
\newblock {\em arXiv preprint arXiv:2309.06419}, 2023.

\bibitem{wu2023exploring}
Zihao Wu, Lu~Zhang, Chao Cao, Xiaowei Yu, Haixing Dai, Chong Ma, Zhengliang Liu, Lin Zhao, Gang Li, Wei Liu, et~al.
\newblock Exploring the trade-offs: Unified large language models vs local fine-tuned models for highly-specific radiology nli task.
\newblock {\em arXiv preprint arXiv:2304.09138}, 2023.

\bibitem{liu2023pharmacygpt}
Zhengliang Liu, Zihao Wu, Mengxuan Hu, Bokai Zhao, Lin Zhao, Tianyi Zhang, Haixing Dai, Xianyan Chen, Ye~Shen, Sheng Li, et~al.
\newblock Pharmacygpt: The ai pharmacist.
\newblock {\em arXiv preprint arXiv:2307.10432}, 2023.

\bibitem{ma2024iterative}
Chong Ma, Zihao Wu, Jiaqi Wang, Shaochen Xu, Yaonai Wei, Zhengliang Liu, Fang Zeng, Xi~Jiang, Lei Guo, Xiaoyan Cai, et~al.
\newblock An iterative optimizing framework for radiology report summarization with chatgpt.
\newblock {\em IEEE Transactions on Artificial Intelligence}, 2024.

\bibitem{lyu2024gp}
Yanjun Lyu, Zihao Wu, Lu~Zhang, Jing Zhang, Yiwei Li, Wei Ruan, Zhengliang Liu, Xiaowei Yu, Chao Cao, Tong Chen, et~al.
\newblock Gp-gpt: Large language model for gene-phenotype mapping.
\newblock {\em arXiv preprint arXiv:2409.09825}, 2024.

\bibitem{zhao2023brain}
Lin Zhao, Lu~Zhang, Zihao Wu, Yuzhong Chen, Haixing Dai, Xiaowei Yu, Zhengliang Liu, Tuo Zhang, Xintao Hu, Xi~Jiang, et~al.
\newblock When brain-inspired ai meets agi.
\newblock {\em Meta-Radiology}, page 100005, 2023.

\bibitem{lewis2020retrieval}
Patrick Lewis, Ethan Perez, Aleksandra Piktus, Fabio Petroni, Vladimir Karpukhin, Naman Goyal, Heinrich K{\"u}ttler, Mike Lewis, Wen-tau Yih, Tim Rockt{\"a}schel, et~al.
\newblock Retrieval-augmented generation for knowledge-intensive nlp tasks.
\newblock {\em Advances in Neural Information Processing Systems}, 33:9459--9474, 2020.

\bibitem{alpaca}
Rohan Taori, Ishaan Gulrajani, Tianyi Zhang, Yann Dubois, Xuechen Li, Carlos Guestrin, Percy Liang, and Tatsunori~B. Hashimoto.
\newblock Stanford alpaca: An instruction-following llama model.
\newblock \url{https://github.com/tatsu-lab/stanford_alpaca}, 2023.

\bibitem{jin2024genegpt}
Qiao Jin, Yifan Yang, Qingyu Chen, and Zhiyong Lu.
\newblock Genegpt: Augmenting large language models with domain tools for improved access to biomedical information.
\newblock {\em Bioinformatics}, 40(2):btae075, 2024.

\bibitem{zakka2024almanac}
Cyril Zakka, Rohan Shad, Akash Chaurasia, Alex~R Dalal, Jennifer~L Kim, Michael Moor, Robyn Fong, Curran Phillips, Kevin Alexander, Euan Ashley, et~al.
\newblock Almanac—retrieval-augmented language models for clinical medicine.
\newblock {\em NEJM AI}, 1(2):AIoa2300068, 2024.

\bibitem{kang2024knowledge}
Minki Kang, Seanie Lee, Jinheon Baek, Kenji Kawaguchi, and Sung~Ju Hwang.
\newblock Knowledge-augmented reasoning distillation for small language models in knowledge-intensive tasks.
\newblock {\em Advances in Neural Information Processing Systems}, 36, 2024.

\bibitem{li2024llava}
Chunyuan Li, Cliff Wong, Sheng Zhang, Naoto Usuyama, Haotian Liu, Jianwei Yang, Tristan Naumann, Hoifung Poon, and Jianfeng Gao.
\newblock Llava-med: Training a large language-and-vision assistant for biomedicine in one day.
\newblock {\em Advances in Neural Information Processing Systems}, 36, 2024.

\bibitem{yao2023react}
Shunyu Yao, Jeffrey Zhao, Dian Yu, Nan Du, Izhak Shafran, Karthik Narasimhan, and Yuan Cao.
\newblock {ReAct}: Synergizing reasoning and acting in language models.
\newblock In {\em International Conference on Learning Representations (ICLR)}, 2023.

\bibitem{shinn2023reflexion}
Noah Shinn, Federico Cassano, Ashwin Gopinath, Karthik Narasimhan, and Shunyu Yao.
\newblock Reflexion: language agents with verbal reinforcement learning.
\newblock In {\em Proceedings of the 37th International Conference on Neural Information Processing Systems}, 2024.

\bibitem{zhang2024surveymemorymechanismlarge}
Zeyu Zhang, Xiaohe Bo, Chen Ma, Rui Li, Xu~Chen, Quanyu Dai, Jieming Zhu, Zhenhua Dong, and Ji-Rong Wen.
\newblock A survey on the memory mechanism of large language model based agents, 2024.

\bibitem{wei2022chain}
Jason Wei, Xuezhi Wang, Dale Schuurmans, Maarten Bosma, Fei Xia, Ed~Chi, Quoc~V Le, Denny Zhou, et~al.
\newblock Chain-of-thought prompting elicits reasoning in large language models.
\newblock {\em Advances in neural information processing systems}, 35:24824--24837, 2022.

\bibitem{singhal2023large}
Karan Singhal, Shekoofeh Azizi, Tao Tu, S~Sara Mahdavi, Jason Wei, Hyung~Won Chung, Nathan Scales, Ajay Tanwani, Heather Cole-Lewis, Stephen Pfohl, et~al.
\newblock Large language models encode clinical knowledge.
\newblock {\em Nature}, 620(7972):172--180, 2023.

\bibitem{luo2022biogpt}
Renqian Luo, Liai Sun, Yingce Xia, Tao Qin, Sheng Zhang, Hoifung Poon, and Tie-Yan Liu.
\newblock Biogpt: generative pre-trained transformer for biomedical text generation and mining.
\newblock {\em Briefings in bioinformatics}, 23(6):bbac409, 2022.

\bibitem{chang2023learning}
Jonathan~D Chang, Kiante Brantley, Rajkumar Ramamurthy, Dipendra Misra, and Wen Sun.
\newblock Learning to generate better than your llm.
\newblock {\em arXiv preprint arXiv:2306.11816}, 2023.

\bibitem{zhong2024evaluationopenaio1opportunities}
Tianyang Zhong, Zhengliang Liu, Yi~Pan, Yutong Zhang, Yifan Zhou, Shizhe Liang, Zihao Wu, Yanjun Lyu, Peng Shu, Xiaowei Yu, Chao Cao, Hanqi Jiang, Hanxu Chen, Yiwei Li, Junhao Chen, Huawen Hu, Yihen Liu, Huaqin Zhao, Shaochen Xu, Haixing Dai, Lin Zhao, Ruidong Zhang, Wei Zhao, Zhenyuan Yang, Jingyuan Chen, Peilong Wang, Wei Ruan, Hui Wang, Huan Zhao, Jing Zhang, Yiming Ren, Shihuan Qin, Tong Chen, Jiaxi Li, Arif~Hassan Zidan, Afrar Jahin, Minheng Chen, Sichen Xia, Jason Holmes, Yan Zhuang, Jiaqi Wang, Bochen Xu, Weiran Xia, Jichao Yu, Kaibo Tang, Yaxuan Yang, Bolun Sun, Tao Yang, Guoyu Lu, Xianqiao Wang, Lilong Chai, He~Li, Jin Lu, Lichao Sun, Xin Zhang, Bao Ge, Xintao Hu, Lian Zhang, Hua Zhou, Lu~Zhang, Shu Zhang, Ninghao Liu, Bei Jiang, Linglong Kong, Zhen Xiang, Yudan Ren, Jun Liu, Xi~Jiang, Yu~Bao, Wei Zhang, Xiang Li, Gang Li, Wei Liu, Dinggang Shen, Andrea Sikora, Xiaoming Zhai, Dajiang Zhu, and Tianming Liu.
\newblock Evaluation of openai o1: Opportunities and challenges of agi, 2024.

\bibitem{kleinpell2021}
Ruth Kleinpell, W.~Grabenkort, W.~Boyle, David Vines, and Keith Olsen.
\newblock The society of critical care medicine at 50 years: Interprofessional practice in critical care looking back and forging ahead.
\newblock {\em Critical Care Medicine}, Publish Ahead of Print, 08 2021.

\bibitem{ervin2018}
Jennifer Ervin, Jeremy Kahn, Taya Cohen, and Laurie Weingart.
\newblock Teamwork in the intensive care unit.
\newblock {\em American Psychologist}, 73:468--477, 05 2018.

\bibitem{chen2023universal}
Xinyun Chen, Renat Aksitov, Uri Alon, Jie Ren, Kefan Xiao, Pengcheng Yin, Sushant Prakash, Charles Sutton, Xuezhi Wang, and Denny Zhou.
\newblock Universal self-consistency for large language model generation.
\newblock {\em arXiv preprint arXiv:2311.17311}, 2023.

\bibitem{podder2023soap}
V.~Podder, V.~Lew, and S.~Ghassemzadeh.
\newblock Soap notes.
\newblock In {\em StatPearls [Internet]}. StatPearls Publishing, Treasure Island (FL), 2023.
\newblock PMID: 29489268, Updated 2024 Jan--.

\bibitem{ely2000}
John Ely, Jerome Osheroff, Mark Ebell, George Bergus, Barcey Levy, M~Chambliss, and Eric Evans.
\newblock Analysis of questions asked by family doctors regarding patient care.
\newblock {\em The Western journal of medicine}, 172:315--9, 06 2000.

\end{thebibliography}

\appendix

\section{A: Agent Interactions}

\subsection{AgentClinic} \label{app:agentclinic}
\subsubsection{MedQA Example \#1} \label{app:agentclinic-all}
\textbf{Interaction with GPT-4 as the backbone for all agents, which produced an incorrect diagnosis:}
\begin{quote}
Doctor: Hello, I'm Dr. Agent. Can you tell me more about the knee pain you're experiencing? When did it start?

Patient: Hello, Dr. Agent. The pain started yesterday. It's been getting worse, especially when I move my knee. I think it might have something to do with playing volleyball recently; there was a lot of jumping and twisting.

Doctor: Can you describe the pain for me? Is it a sharp, stabbing pain, or more of a dull, aching sensation?

Patient: It's more of a dull, aching sensation, but it gets sharp when I try to pivot or twist my knee, like when I'm turning around quickly.

Doctor: Have you noticed any swelling or changes in the appearance of your knee since the pain started?

Patient: No, I haven't noticed any swelling or changes in how my knee looks. It just hurts, especially on the inside part near my kneecap.

Doctor: Have you had any previous injuries to this knee or any surgeries on it in the past?

Patient: No, I haven't had any previous injuries or surgeries on this knee. This is the first time I've had this kind of pain.
Doctor: Have you tried any treatments at home, like applying ice or taking any over-the-counter pain medications? If so, did they provide any relief?

Patient: I've tried icing it a few times since yesterday and took some ibuprofen. It helped a little with the pain, but it's still pretty uncomfortable, especially when I move around or put weight on it.

Doctor: Do you feel any instability in your knee, like it's going to give out on you when you're walking or standing?

Patient: No, I haven't felt like my knee is going to give out on me. It just hurts, but it feels stable enough when I'm walking or standing.

Doctor: Have you experienced any popping or clicking sounds in your knee since the pain started?

Patient: No, I haven't heard any popping or clicking sounds. It's just the pain and discomfort that's been bothering me.

Doctor: Does the pain in your knee worsen with specific movements, such as kneeling, squatting, or going up and down stairs?

Patient: Yes, definitely. Going up and down stairs makes it worse, and squatting is pretty painful too. Kneeling isn't comfortable either.

Doctor: Given your symptoms and the activities that exacerbate your knee pain, it's important to assess the structures within your knee that could be affected. REQUEST TEST: MRI of the knee. This will help us get a detailed view of the soft tissues, including ligaments and cartilage, to better understand the cause of your pain.

Measurement: RESULTS: MRI\_Knee: Optional - Can be considered if clinical suspicion persists despite negative initial workup, not mandatory for diagnosis in this simulated case.

Doctor: Given the information from the MRI suggestion and your symptoms, let's focus on your recent activities. Have you been doing any new exercises or sports, other than volleyball, that you're not accustomed to?

Patient: No, I haven't started any new exercises or sports recently. Volleyball is the only new activity I've been doing, and it's a lot more intense than what I'm used to.

Doctor: Given that volleyball is the only new intense activity you've been doing, it's possible that the pain is related to overuse or strain from the sport. Have you been warming up and stretching properly before playing?

Patient: I try to warm up and stretch a bit before playing, but I might not be doing enough. I usually just jump right into playing without spending much time on warming up or stretching.

Doctor: It's important to properly warm up and stretch before engaging in any physical activity to prevent injuries. Have you noticed if the pain improves with rest or worsens with continued activity?

Patient: The pain does seem to improve a bit with rest. When I sit down and don't move around much, it feels better. But once I start moving again or try to play, it definitely gets worse.

Doctor: Given that rest seems to alleviate the pain somewhat, it's likely that the pain is indeed related to overuse or a strain from playing volleyball. Have you been using any specific type of footwear while playing that might not be providing adequate support?

Patient: Actually, now that you mention it, I've just been using my regular sneakers, not anything specifically designed for volleyball. I didn't think about the impact that might have on my knees.

Doctor: Using appropriate footwear, especially for sports like volleyball that involve a lot of jumping and lateral movements, is crucial to prevent injuries and provide adequate support. Have you noticed any improvement in your knee pain on the days you don't play volleyball?

Patient: Yes, on the days I don't play volleyball, my knee definitely feels better. It seems like the rest really does help.

Doctor: Given that rest improves your symptoms and considering your recent increase in physical activity with inadequate footwear, it sounds like we're dealing with a case of patellar tendinopathy, also known as jumper's knee, which is common in sports with a lot of jumping and landing. Have you been applying ice regularly after playing volleyball, and have you tried any form of knee support, like a brace or a band, while playing?
Patient: I've been applying ice after playing, which seems to help a bit with the pain. I haven't tried using a knee brace or band yet, but that sounds like a good idea. Maybe it could give me some extra support while I'm playing.

Doctor: Considering the positive effect of icing and the potential benefit of additional support, I recommend trying a patellar tendon strap or a knee brace designed for sports. These can help alleviate the strain on your knee during activities. Have you been doing any specific exercises to strengthen the muscles around your knee to support it better?

Patient: No, I haven't started any specific exercises for my knee yet. I wasn't sure what would be safe to do without making the pain worse.

Doctor: Given your situation, it's important to start incorporating some gentle strengthening and flexibility exercises tailored for your knee. This can help support and stabilize the knee joint, potentially reducing pain and preventing future injuries. Would you like me to refer you to a physical therapist who can design a personalized exercise program for you?

Patient: Yes, that sounds like a great idea, Dr. Agent. I'd appreciate a referral to a physical therapist who can help me with a personalized exercise program. I want to make sure I'm doing everything right to get better and avoid any future injuries.

Doctor: DIAGNOSIS READY: Patellar Tendinopathy (Jumper's Knee)

The diagnosis was INCORRECT

\end{quote}

\textbf{Same interaction with o1 as the backbone for all agents, which produced a correct diagnosis:}
\begin{quote}
Doctor: Hello, I'm Dr. Agent. Can you describe when your knee pain began and if you've noticed any swelling or stiffness?

Patient: It started yesterday and has been getting worse since then. The pain is just below my kneecap on the inside of my leg. I haven't noticed any swelling or stiffness.

Doctor: Thank you for sharing that. Could you tell me about any new or increased physical activities you've been doing recently? Also, does the pain get worse during specific movements or activities?

Patient: Yes, I recently joined a volleyball team and have been playing a lot more. The pain does get worse when I'm pivoting or twisting my knee during games. When I'm resting, it doesn't bother me as much.

Doctor: Thank you for the details. Do you remember any specific moment during volleyball when the pain first started, perhaps a twist or awkward landing? Also, have you felt any instability in your knee or heard any popping sounds?

Patient: Actually, I don't remember any specific moment during volleyball when the pain started; it just began after practice. I haven't felt any instability in my knee or heard any popping sounds. It just seems to hurt more with certain movements during the game.

Doctor: Thank you for sharing that information. Have you experienced any pain when climbing stairs or after sitting for a long time? Also, have you ever had any previous knee injuries or issues with your joints?

Patient: Actually, I haven't noticed any pain when climbing stairs or after sitting for a long time. This is the first time I've experienced knee pain, and I've never had any previous injuries or issues with my joints.

Doctor: Thank you for telling me that. I'd like to examine your knee now. Could you please point to the exact spot where you feel the most pain? Also, have you noticed any warmth or redness in that area?

Patient: Yes, the pain is right here. **[Points to the spot just below the kneecap on the inside of the leg.]** I haven't noticed any warmth or redness in that area.

Doctor: Thank you for indicating the exact spot. I'd like to perform some physical tests to assess your knee. Please let me know if you feel any pain or discomfort during these movements.

Patient: Sure, please go ahead. **[During the examination]** Ouch, I feel some pain when you press there on the inside just below my kneecap.

Doctor: Thank you for helping me with the examination. Based on your symptoms, I'd like to order an MRI of your knee to get a closer look at the area. **REQUEST TEST: Knee\_MRI**

Measurement: RESULTS: NORMAL READINGS

Doctor: Your MRI results are normal, which is reassuring. Have you tried applying ice or taking any anti-inflammatory medication, and if so, has it helped? Also, have you noticed any changes in your footwear or playing surface that might contribute to your knee pain?

Patient: No, I haven't tried applying ice or taken any anti-inflammatory medication yet. My shoes and the playing surface have been the same since I started playing, so I haven't noticed any changes that might be causing the pain.

Doctor: Based on your symptoms and examination, I believe you may have pes anserine bursitis, which is an inflammation of the bursa on the inside of your knee below the kneecap. DIAGNOSIS READY: Pes Anserine Bursitis

The diagnosis was CORRECT

\end{quote}

\subsubsection{MedQA Example \#2} \label{app:agentclinic-doctor}
\textbf{Interaction with o1 as the backbone for all agents, which produced an incorrect diagnosis:}
\begin{quote}

Doctor: Hello, I'm Dr. Agent. Could you tell me if you've noticed any unusual symptoms or concerns with your newborn since birth?

Patient: No, doctor, I haven't noticed anything unusual. She's been feeding well and sleeps most of the time.

Doctor: I'm glad to hear that. Has she been gaining weight steadily since birth? Also, have you noticed any yellowing of her skin or eyes, or any unusual rashes?

Patient: Yes, doctor, she's been gaining weight steadily since birth. I haven't noticed any yellowing of her skin or eyes. However, she does have some bluish spots on her skin that don't fade when I press on them.

Doctor: Thank you for telling me about the bluish spots. Could you describe where on her body these spots are located and when you first noticed them?

Patient: I first noticed the spots right after she was born, doctor. They’re on her torso and limbs, and they don't fade when I press on them.

Doctor: Thank you for sharing that. Does anyone in your family have a history of bleeding or clotting disorders? Also, were there any complications during your pregnancy or delivery?

Patient: No, doctor, there's no history of bleeding or clotting disorders in my family that I'm aware of. My pregnancy and delivery went smoothly, although I didn't receive any prenatal care.

Doctor: I understand. To determine the cause of these bluish spots, I'd like to run some blood tests to check her platelet count and clotting function. REQUEST TEST: Complete Blood Count and Coagulation Profile.

Measurement: RESULTS: NORMAL READINGS

Doctor: Thank you for the test results. Since her blood tests are normal, could you tell me about your family's ethnic background? Also, have the bluish spots changed in size or appearance since she was born?

Patient: Well, doctor, we're of Mexican descent. The spots haven't changed in size or appearance since she was born—they look the same as when I first noticed them.

Doctor: Thank you for sharing that information. Based on what you've told me, I believe I have a diagnosis. DIAGNOSIS READY: Mongolian spots (congenital dermal melanocytosis).

The diagnosis was INCORRECT

\end{quote}

\textbf{Same interaction with o1 as the backbone for the doctor agent only while the other three agents used GPT-4 as their backbone, which produced a correct diagnosis:}
\begin{quote}

Doctor: Doctor: Could you please tell me if you've noticed any unusual symptoms in your newborn, such as feeding difficulties, skin discoloration, or breathing problems?

Patient: Yes, I've noticed some bluish spots on her skin that don't go away when I press on them, and her eyes looked a bit cloudy when the light shines on them.

Doctor: Thank you for sharing that. During your pregnancy, did you experience any illnesses like a rash or fever? Also, have you noticed any issues with her feeding or hearing?

Patient: No, I didn't have any illnesses like a rash or fever during pregnancy. But I've noticed she doesn't seem to react to sounds the way I expected. Feeding has been okay, though.

Doctor: Thank you for sharing that. Is there any family history of bleeding disorders, vision problems, or hearing loss? Also, did you take any medications or experience any unusual exposures during your pregnancy?

Patient: No, there's no family history of bleeding disorders, vision problems, or hearing loss that I'm aware of. I didn't take any medications or have any unusual exposures during my pregnancy.

Doctor: Thank you for that information. Just to confirm, did you receive the rubella vaccine before your pregnancy? Also, have you noticed any unusual patterns in her crying or any signs of a heart murmur?

Patient: I'm not sure about the rubella vaccine; I didn't really see a doctor before or during my pregnancy. And yes, there's this constant heart murmur sound that I've noticed; it doesn't seem to stop.

Doctor: Thank you for informing me. To better assess your baby's condition, I'd like to conduct some tests. REQUEST TEST: Echocardiogram, Ophthalmologic Examination, Hearing Assessment, Rubella IgM Serology.

Measurement: RESULTS: - Echocardiogram Findings: Patent ductus arteriosus - Ophthalmologic Consult Findings: Congenital cataracts - Hearing Test Findings: Bilateral sensorineural hearing loss - TORCH Screen: Rubella IgM Positive, Others Negative

Doctor: Based on the test results, it appears that your baby has congenital rubella syndrome. DIAGNOSIS READY: Congenital Rubella Syndrome.

The diagnosis was CORRECT 

\end{quote}

\subsubsection{MedQA Example \#3} \label{app:agentclinic-patient}
\textbf{Interaction with o1 as the backbone for the patient agent only while the other three agents used GPT-4 as their backbone, which produced an incorrect diagnosis:}
\begin{quote}

Doctor: How long have you been experiencing these symptoms of fatigue, flank pain, and fever?

Patient: I've been feeling fatigued with flank pain and a fever for about two weeks now. These symptoms started even before my sinusitis cleared up, and they just haven't gone away.

Doctor: Did the flank pain and fever start at the same time as the fatigue, or did one symptom appear before the others?

Patient: Actually, the fatigue started first, maybe three or four days before I began noticing the flank pain and fever. Then the flank pain and fever appeared around the same time.

Doctor: Have you noticed any changes in your urine, such as color, frequency, or pain during urination?

Patient: No, I haven't noticed any changes in my urine. There's been no pain when I urinate, and the color and frequency seem normal.

Doctor: Have you experienced any nausea, vomiting, or changes in your appetite since these symptoms began?

Patient: No, I haven't experienced any nausea or vomiting, and my appetite has been normal.

Doctor: Have you had any recent injuries or trauma to your back or side where the flank pain is located?

Patient: No, I haven't had any injuries or trauma to my back or side. The pain just started on its own.

Doctor: Have you taken any new medications or supplements, besides the antibiotics for your sinusitis, that could be related to these symptoms?

Patient: No, I haven't started any new medications or supplements besides the antibiotics for my sinusitis. I've only been taking my usual medications—lisinopril, simvastatin, and omeprazole—that I've been on for years.

Doctor: Have you had any previous episodes of kidney stones or urinary tract infections that might explain the flank pain and fever?

Patient: No, I've never had kidney stones or urinary tract infections before. This is the first time I've experienced flank pain and a persistent fever like this.

Doctor: Have you experienced any rash, joint pain, or swelling since these symptoms started?

Patient: No, I haven't experienced any rash since these symptoms started. The rash I had during the antibiotic treatment cleared up after I switched medications. I haven't had any joint pain or swelling, either.

Doctor: Have you had any recent exposure to sick individuals or traveled to areas known for infectious diseases?

Patient: No, I haven't been around anyone who's been sick recently, and I haven't traveled anywhere lately. I've mostly been at home since retiring.

Doctor: Have you experienced any shortness of breath, chest pain, or palpitations since these symptoms started?

Patient: No, I haven't experienced any shortness of breath, chest pain, or palpitations since these symptoms started.

Doctor: Have you noticed any weight loss without trying, night sweats, or a decrease in your overall energy level apart from the fatigue you mentioned?

Patient: No, I haven't noticed any unintentional weight loss or night sweats. Other than the fatigue I've mentioned, my energy levels are about the same.

Doctor: Have you had any changes in your bowel movements or any gastrointestinal symptoms since these symptoms started?

Patient: No, I haven't noticed any changes in my bowel movements or any new gastrointestinal symptoms. I still have my usual gastric reflux, but nothing different or unusual.

Doctor: Have you had any recent blood work or urine tests done since these symptoms started?

Patient: No, I haven't had any blood work or urine tests since these symptoms began. The last time I had any tests done was during my annual check-up a few months ago.

Doctor: Given your symptoms and history, it's important we check for any underlying infections or conditions that might not be immediately obvious. REQUEST TEST: Urinalysis

Measurement: RESULTS: Urinalysis shows Leukocyturia (presence of white blood cells in urine) indicating a possible infection, despite the Bacterial Culture being Negative. Additionally, Cytospin Stained With Hansel’s Solution revealed 3\% binucleated cells with eosinophilic, granular cytoplasm.

Doctor: Given the urinalysis results, it's crucial to explore further. REQUEST TEST: Complete Blood Count (CBC)

Measurement: RESULTS: NORMAL READINGS

Doctor: Given the normal CBC results but abnormal findings in your urinalysis, it's important to consider all possibilities. Have you had any fever spikes or fluctuations in your body temperature in the past few days?

Patient: No, my fever has been pretty steady. I haven't noticed any spikes or fluctuations in my body temperature in the past few days.

Doctor: Given the steady fever and the findings from your urinalysis, it's important to consider your kidney function next. REQUEST TEST: Kidney Function Test

Measurement: RESULTS: Serum Biochemistry shows Elevated Urea and Elevated Creatinine, indicating impaired kidney function.

Doctor: Given the elevated urea and creatinine levels, it's important to assess for any possible sources of infection or obstruction. Have you experienced any increase in the severity of your flank pain or any new symptoms since our last discussion?

Patient: Actually, yes, the flank pain has become more intense over the past few days. Besides that, I haven't noticed any new symptoms.

Doctor: Given the intensification of your flank pain and the test results indicating impaired kidney function, it's crucial to determine if there's any obstruction or specific infection in the kidneys. REQUEST TEST: Renal Ultrasound

Measurement: RESULTS: NORMAL READINGS

Doctor: DIAGNOSIS READY: Acute Pyelonephritis

The diagnosis was INCORRECT

\end{quote}

\textbf{Interaction with GPT-4 as the backbone for all agents, which produced a correct diagnosis:}
\begin{quote}

Doctor: Hello, I'm Dr. Agent. Can you tell me more about the antibiotic treatment you received for your acute sinusitis?

Patient: Yes, of course, Dr. Agent. I was first given amoxicillin for about 12 days, but then I developed a rash. After that, my doctor switched me to cephalexin for another week, and thankfully, the rash went away after the switch.

Doctor: Did you notice any changes in your urine, such as color, frequency, or pain during urination, after starting the antibiotics?

Patient: No, Dr. Agent, I didn't notice any changes in my urine. There was no difference in color, frequency, or any pain during urination after starting the antibiotics.

Doctor: Have you experienced any nausea, vomiting, or changes in your bowel movements since these symptoms started?

Patient: No, Dr. Agent, I haven't experienced any nausea, vomiting, or significant changes in my bowel movements since these symptoms started.

Doctor: Have you had any previous kidney issues or urinary tract infections before this episode?

Patient: No, Dr. Agent, I haven't had any previous kidney issues or urinary tract infections before this.

Doctor: Have you noticed any swelling in your legs or around your eyes since these symptoms started?

Patient: No, Dr. Agent, I haven't noticed any swelling in my legs or around my eyes since these symptoms started.

Doctor: Have you had any recent exposure to new foods, environments, or animals that could have triggered an allergic reaction or infection?

Patient: No, Dr. Agent, I haven't had any recent exposure to new foods, environments, or animals that could have triggered an allergic reaction or infection.

Doctor: Have you experienced any shortness of breath, chest pain, or palpitations since these symptoms started?

Patient: No, Dr. Agent, I haven't experienced any shortness of breath, chest pain, or palpitations since these symptoms started.

Doctor: Have you had any recent weight loss, night sweats, or chills along with your current symptoms?

Patient: Yes, Dr. Agent, along with the fatigue, flank pain, and fever, I've also experienced some night sweats. No significant weight loss, though.

Doctor: Have you noticed any blood in your urine or any unusual discharge?

Patient: No, Dr. Agent, I haven't noticed any blood in my urine or any unusual discharge.

Doctor: Have you been feeling unusually tired or fatigued more than usual, and how has your appetite been since these symptoms started?

Patient: Yes, Dr. Agent, I've been feeling unusually tired and fatigued more than usual. My appetite has decreased a bit since these symptoms started.

Doctor: Have you taken any over-the-counter medications or supplements to help with your symptoms since they started?

Patient: No, Dr. Agent, I haven't taken any over-the-counter medications or supplements to help with my symptoms since they started.

Doctor: Have you had any recent travel history, especially to areas known for infectious diseases, before these symptoms started?

Patient: No, Dr. Agent, I haven't had any recent travel, especially to areas known for infectious diseases, before these symptoms started.

Doctor: Have you experienced any confusion, dizziness, or headaches along with your other symptoms?

Patient: No, Dr. Agent, I haven't experienced any confusion, dizziness, or headaches along with my other symptoms.

Doctor: Have you had any recent contact with someone who was sick or any outbreaks in your community that you're aware of?

Patient: No, Dr. Agent, I haven't had any recent contact with someone who was sick, nor am I aware of any outbreaks in my community.

Doctor: Have you had any previous allergic reactions to medications before this incident with amoxicillin?

Patient: No, Dr. Agent, I haven't had any previous allergic reactions to medications before this incident with amoxicillin. This was the first time something like that happened to me.

Doctor: Given your symptoms and history, it's important to check your kidney function and look for any signs of infection or inflammation. REQUEST TEST: Kidney Function Test (Blood Urea Nitrogen, Creatinine) and Urinalysis.

Measurement: RESULTS: Serum Biochemistry shows Elevated Urea and Elevated Creatinine. Urinalysis indicates Leukocyturia is Present, Bacterial Culture is Negative, and Cytospin Stained With Hansel’s Solution shows 3\% binucleated cells with eosinophilic, granular cytoplasm.

Doctor: Given the elevated urea and creatinine levels along with leukocyturia, it seems there's a kidney issue at play. Have you experienced any recent back injuries or trauma that could have affected your kidneys?

Patient: No, Dr. Agent, I haven't experienced any recent back injuries or trauma that could have affected my kidneys.

Doctor: Given the information and test results, it's crucial to determine if there's an underlying condition causing these kidney issues. Have you had any history of autoimmune diseases in your family, such as lupus or rheumatoid arthritis?

Patient: No, Dr. Agent, there's no history of autoimmune diseases like lupus or rheumatoid arthritis in my family that I'm aware of.

Doctor: Given your symptoms, the lack of a family history of autoimmune diseases, and the test results, it's important to consider all possibilities. Have you been taking any non-prescription drugs, such as NSAIDs, which might affect your kidneys?

Patient: No, Dr. Agent, I haven't been taking any non-prescription drugs like NSAIDs that might affect my kidneys.

Doctor: DIAGNOSIS READY: Acute Interstitial Nephritis (AIN) likely induced by antibiotic treatment.

The diagnosis was CORRECT

\end{quote}
\newpage
\subsection{MedAgents}
\subsubsection{MedQA Example \#1}
\textbf{Interaction with GPT-4 as the backbone for all agents, which produced an incorrect diagnosis:}
\begin{quote}

Question: "A 30-year-old African American woman comes to the physician for the evaluation of a dry cough and chest discomfort for the past 3 days. During this period, the patient has had headaches, muscle aches, joint pain, fever, and chills. Ten days ago, she was hiking with her family in Mississippi. The patient has asthma that is treated with an albuterol inhaler. Her mother has a lung disease treated with methotrexate. The patient has smoked one pack of cigarettes daily for the past 10 years. Her temperature is 38\u00b0C (100.4\u00b0F). Physical examination shows slight wheezes throughout both lung fields. Laboratory studies and urinalysis are positive for polysaccharide antigen. Bronchoalveolar lavage using silver/PAS-staining shows macrophages filled with a dimorphic fungus with septate hyphae. Which of the following is the most likely cause of this patient's symptoms?", "options": "A": "Legionella pneumophila infection", "B": "Aspergillus fumigatus infection", "C": "Pneumocystis pneumonia", "D": "Histoplasma capsulatum infection", "E": "Blastomyces dermatitidis infection"

Experts: Pulmonology", "Infectious Disease", "Microbiology", "Epidemiology", "Pathology"

Pulmonology Expert: Based on the clinical presentation and laboratory findings, the most likely cause of this patient's symptoms is **Histoplasma capsulatum infection**, commonly known as **histoplasmosis**.**Interpretation of the Condition:**1. **Geographical Exposure:** The patient was hiking in Mississippi 10 days ago. Histoplasma capsulatum is endemic in the Ohio and Mississippi River valleys. It thrives in soil contaminated with bird or bat droppings, which are common in these regions.2. **Symptoms:** The patient exhibits respiratory symptoms (dry cough, chest discomfort) and systemic signs (headaches, muscle aches, joint pain, fever, chills) that are consistent with acute pulmonary histoplasmosis.3. **Laboratory Findings:**   - **Positive Polysaccharide Antigen:** Detection of Histoplasma antigen in urine is a sensitive method for diagnosing histoplasmosis.   - **Bronchoalveolar Lavage Findings:** Silver and Periodic acid-Schiff (PAS) staining reveal macrophages filled with a dimorphic fungus exhibiting septate hyphae. Histoplasma capsulatum is a dimorphic fungus that exists as a mold with hyphae in the environment (at cooler temperatures) and as yeast within the host (at body temperature). In tissue samples, it appears as small (2-5 \u00b5m) yeast cells within macrophages.4. **Imaging and Other Tests:** While not provided, chest imaging in histoplasmosis often shows hilar lymphadenopathy and pulmonary infiltrates.**Noteworthy Aspects of the Case:**- **Risk Factors:** Smoking may exacerbate pulmonary symptoms. Although the patient is not immunocompromised, heavy exposure can lead to symptomatic disease even in immunocompetent individuals.- **Differential Diagnosis:** It's important to distinguish histoplasmosis from other conditions like sarcoidosis, especially since the patient's mother is being treated with methotrexate for a lung disease (methotrexate is often used in sarcoidosis). Both conditions can present with similar symptoms and radiographic findings, but histological examination distinguishes them.- **Occupational and Environmental Exposure:** Awareness of environmental exposures is crucial. Activities like hiking in endemic areas increase the risk of inhaling fungal spores.- **Public Health Implication:** Histoplasmosis is not contagious person-to-person but poses a risk to others who may be exposed in the same environment.**Alarming Features:**- **Potential for Dissemination:** Although less common in immunocompetent individuals, histoplasmosis can disseminate, affecting multiple organ systems.- **Respiratory Compromise:** In individuals with pre-existing lung conditions like asthma, fungal infections can lead to significant respiratory distress.- **Misdiagnosis Risk:** Without proper identification, histoplasmosis can be misdiagnosed, leading to inappropriate treatment.**Recommendations:**- **Antifungal Therapy:** Initiation of appropriate antifungal treatment, such as itraconazole, is recommended.- **Further Evaluation:** Monitoring for signs of dissemination, given the systemic symptoms.- **Avoidance of Risk Factors:** Advising the patient on minimizing exposure to endemic areas and soil disruption activities.---**Conclusion:**This patient's symptoms and laboratory findings are characteristic of an acute histoplasmosis infection acquired during recent environmental exposure in an endemic area. Prompt recognition and treatment are essential to prevent complications.

Infectious Disease Expert: The patient is presenting with symptoms and laboratory findings suggestive of **histoplasmosis**, an infection caused by the dimorphic fungus **Histoplasma capsulatum**. This organism is endemic to the Ohio and Mississippi River valleys, which aligns with her recent history of hiking in Mississippi.**Key aspects supporting this diagnosis:**1. **Geographical Exposure:**   - **Mississippi River Valley:** Histoplasma capsulatum is commonly found in soil contaminated with bird or bat droppings in this region. Hiking activities increase exposure risk due to the disturbance of contaminated soil.2. **Clinical Presentation:**   - **Respiratory Symptoms:** Dry cough and chest discomfort are typical initial manifestations as the lungs are the primary site of infection.   - **Systemic Symptoms:** Fever, chills, headaches, muscle aches, and joint pain indicate systemic involvement, which can occur in acute pulmonary histoplasmosis.   - **Wheezing:** Although not classic for histoplasmosis, wheezing may be exacerbated by her underlying asthma.3. **Laboratory Findings:**   - **Positive Polysaccharide Antigen in Serum and Urine:** Detection of histoplasma antigen is a sensitive method for diagnosing disseminated or acute pulmonary histoplasmosis.   - **Bronchoalveolar Lavage Findings:**     - **Silver/PAS Staining:** Highlights fungal organisms. - **Macrophages Filled with Dimorphic Fungus:** Histoplasma capsulatum appears as small yeast forms within macrophages. - **Septate Hyphae:** Indicates the mold phase of a dimorphic fungus when grown at cooler temperatures (in the environment).4. **Dimorphic Fungus Characteristics:**-**Dimorphism:** Histoplasma capsulatum exists as mold (with hyphae) in the environment and yeast (within macrophages) in human tissue. - **Intracellular Location:** The hallmark of histoplasmosis is the presence of yeast forms within macrophages.**Noteworthy and Concerning Aspects:**- **Risk of Dissemination:**  - While typically more severe in immunocompromised individuals, acute pulmonary histoplasmosis can disseminate, especially in high inoculum exposure, leading to more severe illness.- **Underlying Health Conditions:**  - **Asthma:** May complicate her respiratory status and mask the severity of the fungal infection.  - **Smoking History:** Chronic smoking can impair mucociliary clearance and lung defense mechanisms, increasing vulnerability to infections.- **Family History:** - Her mother's lung disease requiring methotrexate suggests a possible familial predisposition to rheumatologic or autoimmune conditions, though methotrexate is also used for certain lung diseases like sarcoidosis. However, there's no direct link to her current infection but may warrant consideration in her overall assessment.- **Occupational and Recreational Exposures:**  - Activities that disturb soil in endemic areas heighten the risk of inhaling Histoplasma spores.**Conclusion:**The most likely cause of this patient's symptoms is an acute pulmonary infection with **Histoplasma capsulatum**. Prompt recognition is essential to manage potential complications, monitor for signs of dissemination, and initiate appropriate antifungal therapy if indicated. Given her symptoms and risk factors, further evaluation, including imaging and possibly consultation with an infectious disease specialist, would be prudent.

Microbiology Expert: Based on the presented clinical scenario, the patient is exhibiting symptoms indicative of a pulmonary fungal infection. The key elements to consider in diagnosing her condition are:1. **Epidemiological Exposure**:   - **Recent Travel**: She was hiking in Mississippi 10 days ago. The Mississippi and Ohio River valleys are endemic areas for **Histoplasma capsulatum**.   - **Potential Environmental Exposure**: Histoplasma is often found in soil enriched with bird or bat droppings, common in cave areas or wooded regions, which are typical hiking environments.2. **Clinical Presentation**:  - **Respiratory Symptoms**: Dry cough and chest discomfort suggest pulmonary involvement.   - **Constitutional Symptoms**: Headaches, muscle aches, joint pain, fever, and chills indicate a systemic infection.  - **Physical Examination**: Slight wheezes throughout both lung fields align with pulmonary infiltration or inflammation.3. **Laboratory Findings**:   - **Positive Polysaccharide Antigen in Urine**: Histoplasma capsulatum releases antigens detectable in urine and serum. The polysaccharide antigen test is a sensitive method for diagnosing disseminated histoplasmosis.   - **Bronchoalveolar Lavage Findings**:     - **Silver/PAS Staining**: Reveals macrophages filled with a dimorphic fungus.     - **Dimorphic Fungus with Septate Hyphae**: Histoplasma capsulatum is a dimorphic fungus (exists as mold in the environment and yeast in human tissue). In tissues, it appears as small yeast forms within macrophages.     - **Intracellular Location**: Histoplasma is known for residing within macrophages, a hallmark of its pathology. 4. **Risk Factors**:   - **Smoking History**: Smoking can impair mucociliary clearance, increasing susceptibility to pulmonary infections.   - **Asthma**: May predispose to or exacerbate respiratory infections.  - **Family History**: While her mother's lung disease treated with methotrexate suggests potential immunosuppression in her mother, it may hint at a genetic predisposition to respiratory conditions, though this is less directly related.5. **Exclusion of Other Pathogens**:   - **Blastomyces dermatitidis**: Also found in the Mississippi River Valley but typically presents with broad-based budding yeast and is extracellular.   - **Coccidioides immitis**: Found in the Southwestern United States, not Mississippi.   - **Cryptococcus neoformans**: Associated with pigeon droppings, presents with encapsulated yeast visible with India ink stain, and commonly affects immunocompromised patients.   - **Aspergillus species**: Shows acute angle branching hyphae but is not dimorphic and doesn't reside within macrophages.**Most Alarming/Noteworthy Aspects**:- **Intracellular Dimorphic Fungus in Macrophages**: Indicates dissemination potential. Histoplasma can spread from the lungs to other organs, especially in immunocompromised individuals.- **Positive Urine Antigen Test**: Suggests a significant fungal burden, raising concern for possible dissemination beyond the lungs.- **Symptom Onset after Environmental Exposure**: The rapid development of symptoms following hiking suggests acute exposure to a high concentration of fungal spores.- **Underlying Asthma and Smoking**: These factors may exacerbate her current condition and complicate recovery.- **Possible Need for Further Evaluation**: Due to the potential for histoplasmosis to become disseminated, especially in patients with risk factors (e.g., smoking, asthma), further assessment and possibly antifungal therapy may be necessary.**Conclusion**:The most likely cause of this patient's symptoms is infection with **Histoplasma capsulatum**, leading to pulmonary histoplasmosis. The identification of intracellular yeast within macrophages on staining, positive polysaccharide antigen in urine, and recent exposure in an endemic area strongly support this diagnosis.

Epidemiology Expert: "Based on the clinical presentation and laboratory findings, the most likely cause of this patient's symptoms is **Histoplasma capsulatum** infection, leading to **pulmonary histoplasmosis**.**Interpretation of the Condition:**1. **Epidemiological Context:**   - **Geographic Location:** The patient was hiking in **Mississippi**, an area endemic for Histoplasma capsulatum, which is prevalent in the Mississippi and Ohio River valleys.   - **Environmental Exposure:** Hiking can expose individuals to soil enriched with bird or bat droppings, common reservoirs for Histoplasma spores.2. **Clinical Presentation:**   - **Respiratory Symptoms:** Dry cough and chest discomfort are hallmark symptoms of acute pulmonary histoplasmosis.  - **Systemic Symptoms:** Fever, chills, headaches, muscle aches, and joint pain indicate a systemic infection.   - **Asthma and Smoking History:** Pre-existing lung conditions like asthma and a history of smoking may predispose the patient to more severe respiratory symptoms.3. **Laboratory Findings:**  - **Positive Polysaccharide Antigen in Urine and Blood:** Detection of Histoplasma antigen in urine is a sensitive method for diagnosing acute histoplasmosis.   - **Bronchoalveolar Lavage (BAL) Findings:**     - **Silver/PAS Staining:** These stains highlight fungal organisms. Histoplasma appears as small, oval yeast cells within macrophages.     - **Macrophages Filled with Dimorphic Fungus:** Histoplasma is a dimorphic fungus, existing as mold (with septate hyphae) in the environment and transforming into yeast within host tissues.    - **Septate Hyphae:** While septate hyphae are characteristic of the mold form, the presence of yeast cells within macrophages is typical in tissue samples.**Most Alarming or Noteworthy Aspects:**1. **Risk of Disseminated Disease:**   - **Potential for Severe Infection:** While histoplasmosis often presents as a mild respiratory illness, there is a risk of dissemination, especially in immunocompromised patients.   - **Early Signs of Dissemination:** The presence of systemic symptoms (fever, chills, joint pains) raises concern for a more extensive infection.2. **Immunosuppression Considerations:**   - **Family History of Lung Disease on Immunosuppressants:** Her mother's lung disease treated with methotrexate (an immunosuppressant) brings attention to possible genetic or environmental predispositions.  - **Patient's Own Risk Factors:** Although not currently on immunosuppressive therapy, her asthma and smoking history might compromise her pulmonary defense mechanisms.3. **Diagnostic Challenges:**   - **Non-Specific Symptoms:** Early histoplasmosis can mimic other respiratory illnesses, potentially delaying diagnosis.  - **Laboratory Confirmation Essential:** Reliance on antigen detection and microscopic examination highlights the need for specific diagnostic tests.4. **Public Health Implications:**   - **Endemic Area Exposure:** This case underscores the importance of awareness about regional endemic fungi and the risks associated with environmental exposures during activities like hiking.   - **Preventive Measures:** Advising at-risk individuals about protective measures can help reduce infection rates.**Conclusion:**The patient's symptoms, exposure history, and laboratory findings are most consistent with pulmonary histoplasmosis caused by Histoplasma capsulatum. Prompt recognition and appropriate antifungal treatment are essential to prevent complications, especially given her potential risk factors for severe disease.",

Pathology Expert: The patient is presenting with symptoms and laboratory findings consistent with histoplasmosis caused by the fungus Histoplasma capsulatum. Interpretation of the Condition: 1. Exposure History: - Geographic Location: The patient was hiking in Mississippi, an area endemic for Histoplasma capsulatum, which is prevalent in the Ohio and Mississippi River valleys. - Environmental Exposure: Hiking increases the likelihood of exposure to soil enriched with bird or bat droppings, where the fungus thrives. 2. Clinical Symptoms: - Respiratory Symptoms: Dry cough and chest discomfort indicate pulmonary involvement. - Constitutional Symptoms: Fever, chills, headaches, muscle aches, and joint pain are systemic symptoms commonly associated with histoplasmosis. 3. Laboratory Findings: - Positive Polysaccharide Antigen in Urine: Histoplasma capsulatum releases polysaccharide antigens detectable in urine and serum tests. - Bronchoalveolar Lavage Results: - Silver/PAS Staining: Revealed macrophages filled with a dimorphic fungus. - Morphology: Presence of septate hyphae suggests the dimorphic nature of Histoplasma capsulatum, which exists as mold (hyphae) in the environment and yeast within the human host. - Intracellular Yeasts: In human tissue, Histoplasma appears as small (2-5 µm) ovoid yeast cells residing within macrophages. 4. Demographic Factors: - Age and Ethnicity: While histoplasmosis can affect individuals of any age and ethnicity, awareness of its prevalence in certain populations helps in considering the diagnosis. Alarming and Noteworthy Aspects: 1. Risk of Disseminated Infection: - Positive Urinary Antigen: Indicates a significant fungal burden, raising concern for dissemination beyond the lungs. - Potential Organs Affected: Disseminated histoplasmosis can involve the liver, spleen, bone marrow, and central nervous system. - Severity in Immunocompromised Hosts: Although the patient is not known to be immunocompromised, her symptoms suggest a more severe infection that warrants prompt attention. 2. Immunocompetent Host with Severe Symptoms: - Smoking History: Smoking one pack daily for 10 years can impair pulmonary defenses, increasing susceptibility to infection. - Asthma: Chronic respiratory conditions may exacerbate symptoms and complicate the clinical picture. 3. Family Medical History: - Mother on Methotrexate: Suggests a familial predisposition to autoimmune conditions or chronic lung diseases, possibly indicating underlying genetic factors affecting immune responses. - Need for Evaluation: Family history may warrant assessment for subtle immunodeficiencies or autoimmune disorders in the patient. 4. Public Health Considerations: - Awareness of Endemic Fungi: Emphasizes the importance of considering regional fungal infections in patients presenting with respiratory symptoms after travel or exposure. - Preventive Measures: Highlights the need for public education on risks associated with activities in endemic areas. 5. Diagnostic Challenges: - Overlap of Symptoms: Histoplasmosis can mimic other respiratory illnesses such as tuberculosis or sarcoidosis, leading to potential misdiagnosis. - Importance of Specific Testing: Reliance on routine tests may delay diagnosis; specialized fungal stains and antigen detection are crucial. Recommendations: - Immediate Initiation of Antifungal Therapy: - Medications: Antifungal agents like itraconazole for mild to moderate cases; amphotericin B may be required for severe or disseminated infections. - Further Investigation: - Immunological Assessment: Evaluate for possible immunodeficiencies. - Imaging Studies: Chest radiography or CT scan to assess the extent of pulmonary involvement. - Monitoring and Follow-up: - Clinical Response: Regular assessment of symptoms and signs to monitor treatment efficacy. - Laboratory Tests: Serial antigen levels to gauge response to therapy. Conclusion: This case underscores the critical role of considering endemic fungal infections in the differential diagnosis of respiratory illnesses, especially with relevant exposure history. The potential for severe and disseminated disease in an ostensibly immunocompetent individual is particularly noteworthy and necessitates prompt and aggressive management to prevent significant morbidity or mortality.

Synthesized report: Question: A 30-year-old African American woman comes to the physician for the evaluation of a dry cough and chest discomfort for the past 3 days. During this period, the patient has had headaches, muscle aches, joint pain, fever, and chills. Ten days ago, she was hiking with her family in Mississippi. The patient has asthma that is treated with an albuterol inhaler. Her mother has a lung disease treated with methotrexate. The patient has smoked one pack of cigarettes daily for the past 10 years. Her temperature is 38°C (100.4°F). Physical examination shows slight wheezes throughout both lung fields. Laboratory studies and urinalysis are positive for polysaccharide antigen. Bronchoalveolar lavage using silver/PAS-staining shows macrophages filled with a dimorphic fungus with septate hyphae. Which of the following is the most likely cause of this patient’s symptoms? Options: ‘A’: ‘Legionella pneumophila infection’, ‘B’: ‘Aspergillus fumigatus infection’, ‘C’: ‘Pneumocystis pneumonia’, ‘D’: ‘Histoplasma capsulatum infection’, ‘E’: ‘Blastomyces dermatitidis infection’ Total Analysis: ** Based on the comprehensive evaluation of the patient’s history, symptoms, laboratory findings, and analyses from multiple domains, the patient is likely suffering from a pulmonary infection caused by Aspergillus fumigatus. — Detailed Synthesis: 1. Epidemiology and Exposure: - Environmental Exposure: Aspergillus spores are ubiquitous in the environment and can be inhaled during outdoor activities like hiking. - Soil and Organic Matter: Disturbance of soil and decaying vegetation can release Aspergillus spores into the air. 2. Clinical Presentation: - Respiratory Symptoms: Dry cough and chest discomfort are common in pulmonary aspergillosis. - Systemic Symptoms: Fever, headaches, muscle aches, joint pain, and chills indicate an acute infection. - Physical Examination: Slight wheezes throughout both lung fields may be exacerbated by her asthma. 3. Laboratory Findings: - Polysaccharide Antigen Positive: Detection of polysaccharide antigens, such as galactomannan, is associated with Aspergillus infection. - Bronchoalveolar Lavage (BAL) Findings: - Silver/PAS staining shows septate hyphae, which are characteristic of Aspergillus species. - Aspergillus forms septate hyphae with acute-angle branching in tissues. - The presence of septate hyphae in human tissue is a hallmark of Aspergillus infection. - Aspergillus is a monomorphic fungus, existing only in the mold form with hyphae. 4. Risk Factors: - Asthma: Underlying lung disease increases susceptibility to Aspergillus infections. - Smoking History: Chronic smoking impairs lung defenses, facilitating fungal colonization and infection. - Immunocompetence: While Aspergillus commonly infects immunocompromised individuals, patients with chronic lung diseases like asthma are also at increased risk. 5. Differential Diagnosis Exclusion: - Histoplasma capsulatum Infection: - Presents as small intracellular yeast forms within macrophages in human tissues, not septate hyphae. - Histoplasma is a dimorphic fungus, but septate hyphae are not seen in human tissues. - Legionella pneumophila Infection: - Causes high fever, hyponatremia, and gastrointestinal symptoms. - Does not produce fungal elements like septate hyphae in tissues. - Pneumocystis Pneumonia: - Occurs mainly in severely immunocompromised patients (e.g., AIDS patients). - Displays disc-shaped yeast forms on methenamine silver stain, not septate hyphae. - Blastomyces dermatitidis Infection: - Appears as broad-based budding yeast in tissues. - Does not form septate hyphae in human tissues. 6. Recommendations: - Antifungal Therapy: Initiate treatment with antifungal agents effective against Aspergillus fumigatus, such as voriconazole. - Further Evaluation: Obtain imaging studies like a chest CT scan to assess for invasive pulmonary aspergillosis. - Monitor for potential complications, such as bronchiectasis or aspergillomas. - Management of Underlying Conditions: Optimize asthma control to improve respiratory function. - Advise smoking cessation to enhance lung health and reduce future infection risk. 7. Conclusion: The synthesis of clinical presentation, risk factors, and laboratory findings supports the diagnosis of a pulmonary infection caused by Aspergillus fumigatus. The presence of septate hyphae on bronchoalveolar lavage staining and a positive polysaccharide antigen test are hallmark features of this infection. Prompt initiation of appropriate antifungal therapy and management of underlying conditions are essential for the patient’s recovery and to prevent potential complications

Vote History: "Pulmonology": "yes","Infectious Disease": "yes", "Microbiology": "no", "Epidemiology": "yes", "Pathology": "yes", "Pathology": "yes","Pulmonology": "yes", "Infectious Disease": "no", "Microbiology": "yes", "Epidemiology": "no","Pathology": "yes"

The predict answer with Synthesized report is INCORECT

\end{quote}

\textbf{Same interaction with o1 as the backbone for all agents, which produced a correct diagnosis:}
\begin{quote}
Question: A 30-year-old African American woman comes to the physician for the evaluation of a dry cough and chest discomfort for the past 3 days. During this period, the patient has had headaches, muscle aches, joint pain, fever, and chills. Ten days ago, she was hiking with her family in Mississippi. The patient has asthma that is treated with an albuterol inhaler. Her mother has a lung disease treated with methotrexate. The patient has smoked one pack of cigarettes daily for the past 10 years. Her temperature is 38. Physical examination shows slight wheezes throughout both lung fields. Laboratory studies and urinalysis are positive for polysaccharide antigen. Bronchoalveolar lavage using silver/PAS-staining shows macrophages filled with a dimorphic fungus with septate hyphae. Which of the following is the most likely cause of this patient's symptoms?", "options": "A": "Legionella pneumophila infection", "B": "Aspergillus fumigatus infection", "C": "Pneumocystis pneumonia", "D": "Histoplasma capsulatum infection", "E": "Blastomyces dermatitidis infection" 

Experts: Pulmonology", "Infectious Diseases", "Immunology", "Mycology", "Travel Medicine"

Pulonology: The patient's symptoms and history suggest that she is suffering from Histoplasmosis, a fungal infection caused by Histoplasma capsulatum. This fungus is endemic in the Mississippi River valley, which aligns with her recent hiking trip in Mississippi. The infection is often contracted by inhaling the fungal spores from soil that has been contaminated with bird or bat droppings. The most alarming aspects of this case include the patient's history of asthma and smoking, both of which can compromise lung function and make her more susceptible to respiratory infections like Histoplasmosis. The presence of fever, chills, muscle aches, joint pain, and headaches suggest systemic involvement, which can occur in more severe cases of Histoplasmosis. The bronchoalveolar lavage findings of macrophages filled with a dimorphic fungus with septate hyphae further confirm the diagnosis. The positive polysaccharide antigen test is also noteworthy, as it is a common method used to diagnose Histoplasmosis. However, it's important to note that this test can cross-react with other fungal infections, so it should be interpreted in the context of the patient's symptoms and exposure history. Lastly, the fact that the patient's mother has a lung disease treated with methotrexate is also of concern..

Infectious Diseases: The patient's symptoms and the results of her laboratory tests suggest that she is suffering from Histoplasmosis, a fungal infection caused by Histoplasma capsulatum. This fungus is endemic in the Mississippi River valley, which aligns with her recent hiking trip in Mississippi. The fungus is dimorphic, existing as a mold in the environment and as a yeast in the body at body temperature. It is often found in soil and material contaminated with bird or bat droppings. The most alarming aspects of this case include the patient's history of asthma and her smoking habit. Both of these factors can compromise lung function and make her more susceptible to respiratory infections, including fungal ones like Histoplasmosis. Furthermore, her mother's lung disease, treated with methotrexate, could suggest a genetic predisposition to lung conditions, although this is not certain. Histoplasmosis often presents with flu-like symptoms, including fever, chills, muscle aches, and a dry cough, as seen in this patient. The presence of the polysaccharide antigen in her urinalysis and the macrophages filled with a dimorphic fungus with septate hyphae in her bronchoalveolar lavage are indicative of a fungal infection. It's also noteworthy that Histoplasmosis can be severe in individuals with compromised immune systems. In this case, the patient's asthma and smoking habit could potentially exacerbate the infection..

Immunology: The patient's symptoms and the results of the laboratory tests suggest that she is suffering from Histoplasmosis, a fungal infection caused by Histoplasma capsulatum. This fungus is endemic in the Mississippi River valley, which aligns with the patient's recent hiking trip in Mississippi. The fungus is dimorphic, existing as a mold in the environment and as a yeast in the body at body temperature. The septate hyphae seen in the bronchoalveolar lavage are characteristic of the mold form, while the macrophages filled with yeast forms are characteristic of the body's response to the infection. The most alarming aspects of this case are the patient's history of asthma and her smoking habit. Both of these factors can compromise lung function and make her more susceptible to respiratory infections, including fungal infections like Histoplasmosis. The fact that her mother has a lung disease treated with methotrexate, an immunosuppressive drug, could also suggest a genetic predisposition to lung disease. The patient's fever, chills, muscle aches, joint pain, and headaches are systemic symptoms that suggest the infection may have spread beyond the lungs. This is a serious condition that requires immediate treatment. The presence of polysaccharide antigen in the urinalysis is also noteworthy, as it is a marker of Histoplasma infection..

Mycology: The patient's symptoms and the results of the laboratory tests suggest that she is suffering from Histoplasmosis, a fungal infection caused by Histoplasma capsulatum. This fungus is endemic in the Mississippi River valley, which aligns with the patient's recent hiking trip in Mississippi. Histoplasma capsulatum is a dimorphic fungus that can exist in a yeast form at body temperature and a mold form at room temperature. The fungus is often found in soil and material contaminated with bird or bat droppings.The most alarming aspects of this case include the patient's history of asthma and smoking, which could potentially exacerbate her respiratory symptoms and complicate her recovery. The presence of fever and chills also suggests a systemic infection, which could be serious if not treated promptly. Furthermore, the patient's mother's lung disease treated with methotrexate, an immunosuppressant, could potentially indicate a genetic predisposition to respiratory illnesses, although this is speculative and would require further investigation. The detection of the fungus in macrophages filled with septate hyphae is also noteworthy. This is a characteristic feature of Histoplasma capsulatum, which primarily infects macrophages and can survive and multiply within these cells. This ability to evade the immune system makes Histoplasma capsulatum a particularly challenging pathogen to treat.

Travel Medicine: The patient's symptoms and the results of her laboratory tests suggest that she is suffering from Histoplasmosis, a fungal infection caused by Histoplasma capsulatum. This fungus is endemic in the Mississippi River valley, which aligns with her recent hiking trip in Mississippi. The fungus is dimorphic, existing as a mold in the environment and as a yeast in the body. It is often found in soil and material contaminated with bird or bat droppings. The most alarming aspects of this case are the patient's symptoms of fever, chills, dry cough, and chest discomfort, which are indicative of a systemic infection. The presence of the fungus in her bronchoalveolar lavage, as evidenced by the macrophages filled with a dimorphic fungus with septate hyphae, confirms this. Another noteworthy aspect is her history of asthma and smoking, which could have made her more susceptible to respiratory infections. Her mother's lung disease, treated with methotrexate, could also be a concern as some lung diseases have a genetic component. However, it's not clear from the information given whether this is relevant in this case. The positive polysaccharide antigen test is also significant. This test is often used to detect certain types of fungal infections, including Histoplasmosis..

Synthesized report: "Question: A 30-year-old African American woman comes to the physician for the evaluation of a dry cough and chest discomfort for the past 3 days. During this period, the patient has had headaches, muscle aches, joint pain, fever, and chills. Ten days ago, she was hiking with her family in Mississippi. The patient has asthma that is treated with an albuterol inhaler. Her mother has a lung disease treated with methotrexate. The patient has smoked one pack of cigarettes daily for the past 10 years. Her temperature is 38\u00b0C (100.4\u00b0F). Physical examination shows slight wheezes throughout both lung fields. Laboratory studies and urinalysis are positive for polysaccharide antigen. Bronchoalveolar lavage using silver/PAS-staining shows macrophages filled with a dimorphic fungus with septate hyphae. Which of the following is the most likely cause of this patient's symptoms? Options: {'A': 'Legionella pneumophila infection', 'B': 'Aspergillus fumigatus infection', 'C': 'Pneumocystis pneumonia', 'D': 'Histoplasma capsulatum infection', 'E': 'Blastomyces dermatitidis infection'} Total Analysis: [The patient's symptoms, history, and test results strongly suggest a diagnosis of Histoplasmosis, a fungal infection caused by Histoplasma capsulatum. This fungus is endemic in the Mississippi River valley, aligning with her recent hiking trip in Mississippi. The fungus is dimorphic, existing as a mold in the environment and as a yeast in the body at body temperature. It is often found in soil and material contaminated with bird or bat droppings. The patient's history of asthma and smoking, both of which can compromise lung function, make her more susceptible to respiratory infections like Histoplasmosis. The presence of fever, chills, muscle aches, joint pain, and headaches suggest systemic involvement, which can occur in more severe cases of Histoplasmosis. The bronchoalveolar lavage findings of macrophages filled with a dimorphic fungus with septate hyphae further confirm the diagnosis. The positive polysaccharide antigen test is also noteworthy, as it is a common method used to diagnose Histoplasmosis. However, it's important to note that this test can cross-react with other fungal infections, so it should be interpreted in the context of the patient's symptoms and exposure history. Lastly, the fact that the patient's mother has a lung disease treated with methotrexate is also of concern, as it could suggest a genetic predisposition to lung conditions, although this is not certain. Immediate treatment is required to manage this serious condition.

Vote History: Pulmonology": "yes", "Infectious Diseases": "yes", "Immunology": "yes", "Mycology": "yes", "Travel Medicine": "yes"

The predict answer with Synthesized report is CORRECT

\end{quote}

\subsubsection{MedQA Example \#2}

\begin{quote}
Question: A 62-year-old patient has been hospitalized for a week due to a stroke. One week into the hospitalization, he develops a fever and purulent cough. His vitals include: heart rate 88/min, respiratory rate 20/min, temperature 38.4\u00b0C (101.1\u00b0F), and blood pressure 110/85 mm Hg. On physical examination, he has basal crackles on the right side of the chest. Chest radiography shows a new consolidation on the same side. Complete blood count is as follows: Hemoglobin 16 mg/dL Hematocrit 50%\nLeukocyte count 8,900/mm3\nNeutrophils 72%\nBands 4%\nEosinophils 2%\nBasophils 0%\nLymphocytes 17%\nMonocytes 5%\nPlatelet count 280,000/mm3\nWhat is the most likely causal microorganism?",
    "options": {
        "A": "Pseudomona aeruginosa",
        "B": "Streptococcus pneumoniae",
        "C": "Mycobacterium tuberculosis",
        "D": "Haemophilus influenzae",
        "E": "Staphylococcus aureus"
    }

Experts: "Infectious Disease","Pulmonology","Neurology","Microbiology", "Radiology"

Infectious Disease Expert: “The 62-year-old patient is exhibiting signs of a hospital-acquired pneumonia, most likely aspiration pneumonia. The most probable causal microorganisms are anaerobic bacteria from the oropharyngeal flora, such as Bacteroides species, Fusobacterium, and Peptostreptococcus. Interpretation of the Condition: 1. Hospitalization and Stroke History: The patient has been hospitalized for a week due to a stroke. Stroke patients often have impaired consciousness and dysphagia (difficulty swallowing), which increases the risk of aspiration of oropharyngeal contents into the lungs. 2. Symptom Development: One week into hospitalization, the patient develops a fever (38.4°C/101.1°F) and a purulent cough. These are classic signs of a lung infection. 3. Physical Examination: Basal crackles on the right side of the chest indicate the presence of fluid or consolidation in the lung tissue, which is consistent with pneumonia. 4. Chest Radiography: A new consolidation on the right side supports the diagnosis of pneumonia. The right main bronchus is more vertical and wider, making the right lower lobe a common site for aspiration pneumonia. 5. Laboratory Findings: The complete blood count shows a normal leukocyte count but a shift towards neutrophilia (72\% neutrophils and \% bands), which is typical in bacterial infections. The hemoglobin and hematocrit levels are elevated, possibly due to hemoconcentration from dehydration. Noteworthy Aspects: - Risk of Aspiration Pneumonia: The most alarming aspect is the high risk of aspiration pneumonia due to the patient’s stroke. Dysphagia can lead to aspiration of bacteria-laden oropharyngeal secretions. - Hospital-Acquired Infection: Developing pneumonia during hospitalization raises concerns for hospital-acquired (nosocomial) infections, which can be more severe and caused by resistant organisms. - Right Lower Lobe Involvement: The localization of the consolidation to the right lower lobe is characteristic of aspiration pneumonia, emphasizing the need to address swallowing difficulties and prevent further aspiration. - Potential for Anaerobic Infection: Anaerobic bacteria are not covered by all standard antibiotic regimens. Failure to cover anaerobes could lead to inadequate treatment and complications like lung abscesses. Conclusion: Given the clinical scenario, the most likely causative agents are anaerobic bacteria from the oropharyngeal flora due to aspiration during swallowing difficulties associated with his stroke. Prompt recognition and initiation of appropriate antibiotic therapy covering anaerobes are crucial to manage the infection and prevent complications.”

Pulmonology Expert: “The patient is a 62-year-old man who, after being hospitalized for a stroke, develops signs of pneumonia including fever, purulent cough, right-sided basal crackles, and a new right-sided consolidation on chest X-ray. His complete blood count is relatively unremarkable with a normal white blood cell count and differential. Interpretation of the Condition: This clinical picture is highly suggestive of aspiration pneumonia. Patients who have suffered a stroke are at increased risk of aspiration due to potential dysphagia, impaired cough reflex, and altered consciousness. The right lung, particularly the lower lobe, is more commonly affected in aspiration events because the right main bronchus is more vertical and wider than the left, making it the path of least resistance for aspirated material. Most Likely Causal Microorganism: In cases of aspiration pneumonia, the most common causative agents are anaerobic bacteria that are part of the normal oropharyngeal flora. These organisms include species such as Bacteroides, Fusobacterium, Peptostreptococcus, and Prevotella. Noteworthy Aspects: 1. Risk Factors for Aspiration: - The patient’s recent stroke is significant because it likely impairs his swallowing mechanism and cough reflex, increasing the risk of aspiration of oropharyngeal contents. 2. Clinical Signs of Aspiration Pneumonia: - The development of a fever and purulent cough one week into hospitalization aligns with the typical timeline for aspiration pneumonia. - The physical examination finding of basal crackles and the chest radiograph showing a new consolidation on the right side further support the diagnosis. 3. Laboratory Findings: - The normal white blood cell count and differential do not rule out pneumonia. In aspiration pneumonia caused by anaerobes, the inflammatory response may not always result in a significant leukocytosis. - The elevated hemoglobin and hematocrit levels may suggest hemoconcentration, possibly due to dehydration, which can be a concern in hospitalized patients and may exacerbate the risk of thrombotic events, especially in a patient who has recently had a stroke. Conclusion: The most likely causal microorganisms are anaerobic bacteria from the oropharyngeal flora, given the patient’s risk factors and clinical presentation consistent with aspiration pneumonia."

Neurology Expert: “Interpretation of the Condition: The patient is a 62-year-old man who has been hospitalized for a week due to a stroke. One week into his hospitalization, he develops a fever (38.4°C or 101.1°F) and a purulent cough. Physical examination reveals basal crackles on the right side of the chest, and chest radiography shows a new consolidation on the same side. His complete blood count (CBC) shows a normal leukocyte count with a slight left shift (elevated neutrophils and bands), indicating an early bacterial infection. Given the context, the most plausible diagnosis is aspiration pneumonia resulting from the patient’s stroke-related complications. Pathophysiological Explanation: - Stroke and Aspiration Risk: Stroke patients often have neurological deficits that can impair their swallowing reflex (dysphagia) and cough reflex, increasing the risk of aspiration. Aspiration occurs when oropharyngeal or gastric contents enter the lower respiratory tract. - Aspiration Pneumonia: This condition is an infection of the lungs following the inhalation of oropharyngeal secretions colonized by pathogenic bacteria. The right lower lobe is more commonly affected due to the more vertical orientation of the right main bronchus. Most Likely Causal Microorganism: The most likely causative agents in aspiration pneumonia are anaerobic bacteria from the oral flora, including: - Bacteroides fragilis - Prevotella melaninogenica - Fusobacterium nucleatum - Peptostreptococcus species However, in hospitalized patients, particularly those who have been in the hospital for more than 5 days, the spectrum of pathogens shifts towards Gram-negative bacilli and Staphylococcus aureus due to hospital-acquired flora colonization. Considering the patient’s hospital stay and his risk factors, Klebsiella pneumoniae, a Gram-negative bacillus, emerges as the most likely causative organism. Rationale: - Klebsiella pneumoniae Characteristics: - Commonly associated with aspiration pneumonia in elderly and debilitated patients. - Known to cause lobar pneumonia with prominent consolidation. - Often produces a thick, mucoid, and sometimes blood-tinged sputum (“currant jelly” sputum), though this is not always present. - Risk Factors Present: - Age: Patients over 60 are at higher risk. - Hospitalization: A week-long stay increases exposure to hospital-acquired pathogens. - Neurological Deficit: Stroke can impair protective airway reflexes. - Aspiration Risk: Likely due to dysphagia associated with stroke. Alarming and Noteworthy Aspects: 1. Development of Hospital-Acquired Pneumonia (HAP): - The onset of pneumonia after 48 hours of hospitalization classifies it as HAP, which is associated with higher morbidity and mortality rates. - HAP pathogens are often resistant to common antibiotics, complicating treatment. 2. Risk of Multidrug-Resistant Organisms: - Klebsiella pneumoniae in the hospital setting can be resistant to multiple antibiotics, including carbapenems (carbapenem-resistant Enterobacteriaceae). - This resistance necessitates the use of broader-spectrum or combination antibiotic therapies, which have more side effects and risks. 3. Impaired Airway Protection Due to Stroke: - The patient’s stroke may have led to bulbar palsy, affecting swallowing and gag reflexes. - This increases the chance of recurrent aspiration and persistent infection. 4. Potential for Rapid Clinical Deterioration: - Pneumonia in post-stroke patients can quickly lead to respiratory failure. - Close monitoring is required to manage potential complications such as sepsis or acute respiratory distress syndrome (ARDS). 5. Normal WBC Count Despite Infection: - The patient’s leukocyte count is within normal limits, which can be misleading.

“Microbiology”: “Interpretation of the Condition: The 62-year-old patient, who has been hospitalized for a week due to a stroke, is now presenting with signs of a lower respiratory tract infection. The key clinical features include: - Fever and Purulent Cough: Indicative of an infectious process in the lungs. - Basal Crackles on the Right Side: Suggestive of fluid or consolidation in the lower part of the right lung. - Chest Radiography Showing New Consolidation on the Right Side: Confirms the presence of pneumonia in the right lung. The patient’s vital signs show a mild fever (38.4°C), normal heart rate (88/min), slightly elevated respiratory rate (20/min), and normal blood pressure (110/85 mm Hg). The complete blood count (CBC) reveals a leukocyte count within normal limits but with a slight left shift (neutrophils at 72\% and bands at 4\%), which is often seen in bacterial infections. Most Noteworthy Aspects: 1. Recent Stroke and Hospitalization: The patient’s recent stroke is significant because it may have led to dysphagia (difficulty swallowing) and reduced consciousness levels, increasing the risk of aspiration pneumonia. 2. Aspiration Risk Factors: Stroke patients often have impaired swallowing reflexes, making them susceptible to aspirating oropharyngeal or gastric contents into the lungs, particularly affecting the right lung due to the anatomical structure of the bronchial tree. 3. Hospital Environment: After a week in the hospital, the patient is at risk of acquiring nosocomial (hospital-acquired) infections, which are often caused by organisms resistant to common antibiotics. 4. Pathogen Considerations: - Gram-Negative Bacteria: Organisms like Pseudomonas aeruginosa are common causes of hospital-acquired pneumonia. - Anaerobic Bacteria: Aspiration can introduce anaerobic bacteria from the oropharyngeal flora into the lungs. - Staphylococcus aureus: Including MRSA strains, which are prevalent in hospital settings. Most Likely Causal Microorganism: Given the patient’s risk factors (recent stroke leading to potential aspiration, hospitalization duration, and new onset pneumonia in a hospital setting), the most likely causal microorganism is Pseudomonas aeruginosa. — Rationale: Pseudomonas aeruginosa is a common cause of hospital-acquired pneumonia, especially in patients with prolonged hospital stays and those who are debilitated. It is a Gram-negative, opportunistic pathogen known for its resistance to multiple antibiotics and its ability to thrive in hospital environments. The right-sided pneumonia aligns with the likelihood of aspiration in stroke patients, and Pseudomonas is frequently implicated in such cases within hospitals.”,

Radiology Expert: “Based on the presented scenario, the 62-year-old patient, who has been hospitalized for a week due to a stroke, is exhibiting signs of pulmonary infection. The key findings and interpretations are as follows: Clinical Presentation: - Fever and Purulent Cough: These are classic symptoms of a lower respiratory tract infection. - Vital Signs: - Temperature: Elevated at 38.4°C (101.1°F), indicating fever. - Heart Rate and Blood Pressure: Within normal limits. - Respiratory Rate: Slightly elevated at 20/min, suggesting increased respiratory effort. - Physical Examination: Basal crackles on the right side of the chest, which indicate fluid or consolidation in the lung tissue at the base of the lung. Radiological Findings: - Chest Radiography: Shows a new consolidation on the right side, consistent with pneumonia. In post-stroke patients, aspiration commonly affects the right lower lobe because the right main bronchus is more vertical and wider than the left. Laboratory Results: - Complete Blood Count (CBC): - Leukocyte Count: Normal at 8,900/mm³, but with a neutrophilic predominance (72\% neutrophils and 4\% bands), which is indicative of a bacterial infection. - Hemoglobin and Hematocrit: Elevated (16 mg/dL and 50\%, respectively), which may suggest hemoconcentration due to dehydration. - Other Cell Lines: Within normal limits. Interpretation of the Condition: The patient is likely experiencing aspiration pneumonia. Post-stroke patients are at high risk for aspiration due to: - Impaired Swallowing (Dysphagia): Stroke can affect the muscles involved in swallowing, leading to inadvertent passage of oropharyngeal contents into the lower airways. - Depressed Cough Reflex: Weakness or impairment prevents the clearance of aspirated material. - Altered Consciousness: Though not explicitly mentioned, stroke patients may have fluctuating levels of consciousness, increasing aspiration risk. Most Likely Causal Microorganism: Anaerobic Bacteria from the Oral Flora, such as: - Bacteroides species - Prevotella melaninogenica - Fusobacterium nucleatum - Peptostreptococcus species These anaerobes are commonly involved in aspiration pneumonia because they reside in the oropharynx and can be aspirated into the lungs. The anaerobic environment within the consolidated lung tissue allows these bacteria to thrive. Aspects of Concern: 1. Risk of Complications: - Lung Abscess Formation: Anaerobic infections can lead to necrosis and abscess formation within the lung parenchyma. - Empyema: Accumulation of pus in the pleural space. - Sepsis: The infection can become systemic if not promptly treated. 2. Delayed Diagnosis: - Normal Leukocyte Count: May mask the severity of infection; reliance solely on leukocyte count can delay recognition and treatment. - Subtle Clinical Signs: The patient’s vital signs are relatively stable, which may lead to underestimation of the infection’s severity. 3. Swallowing Assessment and Preventive Measures: - Need for Swallowing Evaluation: A speech and swallow assessment is critical to prevent further aspiration events. - Feeding Modifications: Implementation of dietary changes or feeding assistance to reduce aspiration risk. 4. Antibiotic Management: - Necessity for Prompt and Appropriate Antibiotic Therapy: Coverage should include anaerobic organisms. - Antibiotic Resistance: Consideration of local antibiograms is essential to select effective antibiotics. Conclusion: The patient’s presentation is most consistent with aspiration pneumonia caused by anaerobic bacteria from the oral flora. This condition requires immediate medical attention to prevent serious complications. Interdisciplinary management involving neurology, pulmonology, radiology, and speech therapy is crucial to address both the infection and the underlying risk factors contributing to aspiration.”

Synthesized Report: “Question: A 62-year-old patient has been hospitalized for a week due to a stroke. One week into the hospitalization, he develops a fever and purulent cough. His vitals include: heart rate 88/min, respiratory rate 20/min, temperature 38.4°C (101.1°F), and blood pressure 110/85 mm Hg. On physical examination, he has basal crackles on the right side of the chest. Chest radiography shows a new consolidation on the same side. Complete blood count is as follows: Hemoglobin 16 mg/dL Hematocrit 50\% Leukocyte count 8,900/mm3 Neutrophils 72\% Bands 4\% Eosinophils 2\% Basophils 0\% Lymphocytes 17\% Monocytes 5\% Platelet count 280,000/mm3 What is the most likely causal microorganism? Options: {‘A’: ‘Pseudomona aeruginosa’, ‘B’: ‘Streptococcus pneumoniae’, ‘C’: ‘Mycobacterium tuberculosis’, ‘D’: ‘Haemophilus influenzae’, ‘E’: ‘Staphylococcus aureus’} Total Analysis: Considering the clinical scenario and synthesizing the expert analyses, the most likely causative microorganism is Staphylococcus aureus. The patient developed pneumonia one week into hospitalization for a stroke, indicating a hospital-acquired infection. While aspiration pneumonia commonly involves anaerobic bacteria, these are not listed among the provided options. Staphylococcus aureus is a common cause of hospital-acquired pneumonia, especially in patients with risk factors such as impaired swallowing from a stroke leading to aspiration. Although Pseudomonas aeruginosa is also a hospital-acquired pathogen, it is more commonly associated with mechanical ventilation and severe immunosuppression, which are not present in this patient. Therefore, given the patient’s risk factors and the options provided, Staphylococcus aureus is the most probable causative agent.”

Vote History:"Infectious Disease": "yes",
            "Pulmonology": "yes",
            "Neurology": "yes",
            "Microbiology": "yes",
            "Radiology": "yes"

\end{quote}

\begin{quote}
Question: A 62-year-old patient has been hospitalized for a week due to a stroke. One week into the hospitalization, he develops a fever and purulent cough. His vitals include: heart rate 88/min, respiratory rate 20/min, temperature 38.4\u00b0C (101.1\u00b0F), and blood pressure 110/85 mm Hg. On physical examination, he has basal crackles on the right side of the chest. Chest radiography shows a new consolidation on the same side. Complete blood count is as follows:Hemoglobin 16 mg/dLHematocrit 50\%Leukocyte count 8,900/mm3Neutrophils 72\% Bands 4\% Eosinophils 2\% Basophils 0\%Lymphocytes 17\% Monocytes 5\%Platelet count 280,000/mm3 What is the most likely causal microorganism?", "options": "A": "Pseudomona aeruginosa", "B": "Streptococcus pneumoniae", "C": "Mycobacterium tuberculosis", "D": "Haemophilus influenzae", "E": "Staphylococcus aureus"

Experts: "Neurology", "Infectious Disease", "Pulmonology", "Radiology", "Hematology"

Neurology Expert: The patient's symptoms and clinical findings suggest that he has developed hospital-acquired pneumonia (HAP), also known as nosocomial pneumonia. This is a type of pneumonia that occurs 48 hours or more after admission, which was not incubating at the time of admission. The patient's fever, purulent cough, and new consolidation on chest radiography are indicative of this condition.Hospital-acquired pneumonia is most commonly caused by bacteria. The most common pathogens include Staphylococcus aureus, Klebsiella pneumoniae, Escherichia coli, and Pseudomonas aeruginosa. Given the patient's age and the fact that he has been hospitalized for a week, the most likely causal microorganism is Staphylococcus aureus, including methicillin-resistant Staphylococcus aureus (MRSA), or gram-negative bacteria like Klebsiella or Pseudomonas.The most alarming aspect of this case is the development of a hospital-acquired infection, which can complicate the patient's recovery from stroke and can lead to severe complications, including respiratory failure, sepsis, and even death. It's also noteworthy that the patient's leukocyte count is within normal limits, which may suggest that the infection is still in its early stages or that the patient's immune response is not robust. Prompt diagnosis and treatment are crucial in these cases."

Infectious Disease: The patient's symptoms and clinical findings suggest that he has developed hospital-acquired pneumonia (HAP), also known as nosocomial pneumonia. This is a type of pneumonia that occurs 48 hours or more after admission to the hospital. The fever, purulent cough, basal crackles on the right side of the chest, and new consolidation on chest radiography are all indicative of this condition. Hospital-acquired pneumonia is often caused by bacteria that are more resistant to antibiotics than community-acquired pneumonia. The most common pathogens include Staphylococcus aureus (including MRSA), Klebsiella pneumoniae, Pseudomonas aeruginosa, and Escherichia coli. The alarming aspect of this case is the development of pneumonia in a patient who is already hospitalized due to a stroke. This can complicate the patient's recovery and increase the risk of mortality. The patient's age and the fact that he has had a stroke may also make him more susceptible to severe infection. The complete blood count does not show a significant increase in leukocyte count, which is usually seen in bacterial infections. However, this does not rule out a bacterial infection as the cause of the pneumonia. The patient's symptoms and the findings on physical examination and chest radiography are more indicative of the diagnosis. The exact causal microorganism can only be determined by a sputum culture or other appropriate diagnostic tests.."

Pulmonology: The patient's symptoms and clinical findings suggest that he has developed hospital-acquired pneumonia (HAP), also known as nosocomial pneumonia. This is a type of pneumonia that occurs 48 hours or more after admission, which was not incubating at the time of admission. The symptoms include fever, purulent cough, and new lung infiltrate seen on chest radiography. The physical examination finding of basal crackles on the right side of the chest and the radiographic evidence of consolidation on the same side further support this diagnosis.Hospital-acquired pneumonia is often caused by bacteria that are more resistant to antibiotics than community-acquired pneumonia. The most common pathogens include Staphylococcus aureus, Klebsiella pneumoniae, Pseudomonas aeruginosa, and Escherichia coli. Given the patient's clinical scenario, it's not possible to definitively identify the causal microorganism without further diagnostic tests such as sputum culture or blood culture.The most alarming aspect of this case is the development of pneumonia in a patient who is already hospitalized for a stroke. This can complicate the patient's recovery and increase the risk of mortality. It's important to promptly identify the causal microorganism and initiate appropriate antibiotic therapy. Additionally, preventive measures should be taken to reduce the risk of hospital-acquired infections.", "Radiology": "The patient's symptoms and the results of the chest radiography suggest that he has developed hospital-acquired pneumonia (HAP), also known as nosocomial pneumonia. This is a type of pneumonia that occurs 48 hours or more after admission, which was not incubating at the time of admission. The fever, purulent cough, and new consolidation on the chest radiography are typical signs of this condition.Hospital-acquired pneumonia is often caused by bacteria that are more resistant to antibiotics than community-acquired pneumonia. The most common organisms causing HAP are Staphylococcus aureus, Klebsiella pneumoniae, E. coli, and Pseudomonas aeruginosa. Given the patient's hospitalization and the symptoms presented, the most likely causal microorganism is one of these.The alarming aspect of this case is the development of pneumonia in a patient already hospitalized for a stroke. This can complicate the patient's recovery and increase the risk of mortality. The patient's age also puts him at a higher risk of severe complications. Prompt diagnosis and treatment are crucial in such cases.

Hematology Expert: The patient's symptoms and clinical findings suggest a case of hospital-acquired pneumonia (HAP), also known as nosocomial pneumonia. This is a type of pneumonia that occurs 48 hours or more after admission, which was not incubating at the time of admission. The patient's fever, purulent cough, basal crackles on the right side of the chest, and new consolidation on chest radiography are all indicative of this condition.Hospital-acquired pneumonia is often caused by bacteria that are common in the hospital environment. The most common organisms include Staphylococcus aureus (including Methicillin-resistant Staphylococcus aureus or MRSA), Klebsiella pneumoniae, Pseudomonas aeruginosa, and Escherichia coli. The complete blood count does not show any significant abnormalities that would suggest a specific organism. The leukocyte count is slightly elevated, which is common in infections, but it does not provide a clue to the specific organism causing the infection. The most alarming aspect of this case is the development of pneumonia in a patient who is already hospitalized for a stroke. This can complicate the patient's recovery and increase the risk of morbidity and mortality. It's also noteworthy that hospital-acquired infections can be caused by organisms that are resistant to common antibiotics, which can make treatment more challenging.

Radiology Expert: The patient's symptoms and the results of the chest radiography suggest that he has developed hospital-acquired pneumonia (HAP), also known as nosocomial pneumonia. This is a type of pneumonia that occurs 48 hours or more after admission, which was not incubating at the time of admission. The fever, purulent cough, and new consolidation on the chest radiography are typical signs of this condition. Hospital-acquired pneumonia is often caused by bacteria that are more resistant to antibiotics than community-acquired pneumonia. The most common organisms causing HAP are Staphylococcus aureus, Klebsiella pneumoniae, E. coli, and Pseudomonas aeruginosa. Given the patient's hospitalization and the symptoms presented, the most likely causal microorganism is one of these. The alarming aspect of this case is the development of pneumonia in a patient already hospitalized for a stroke. This can complicate the patient's recovery and increase the risk of mortality. The patient's age also puts him at a higher risk of severe complications. Prompt diagnosis and treatment are crucial in such cases.

Synthesized Report: "Question: A 62-year-old patient has been hospitalized for a week due to a stroke. One week into the hospitalization, he develops a fever and purulent cough. His vitals include: heart rate 88/min, respiratory rate 20/min, temperature 38.4\u00b0C (101.1\u00b0F), and blood pressure 110/85 mm Hg. On physical examination, he has basal crackles on the right side of the chest. Chest radiography shows a new consolidation on the same side. Complete blood count is as follows:Hemoglobin 16 mg/dLHematocrit 50\%Leukocyte count 8,900/mm3Neutrophils 72\%Bands 4\%Eosinophils 2\%Basophils \%Lymphocytes 17\%Monocytes 5\%Platelet count 280,000/mm3What is the most likely causal microorganism? Options: {'A': 'Pseudomona aeruginosa', 'B': 'Streptococcus pneumoniae', 'C': 'Mycobacterium tuberculosis', 'D': 'Haemophilus influenzae', 'E': 'Staphylococcus aureus'} Total Analysis: The patient's symptoms, clinical findings, and hospitalization history suggest a diagnosis of hospital-acquired pneumonia (HAP). The most common pathogens causing HAP are Staphylococcus aureus (including MRSA), Klebsiella pneumoniae, Pseudomonas aeruginosa, and Escherichia coli. Given the patient's clinical scenario, the most plausible causal microorganisms are Pseudomonas aeruginosa and Staphylococcus aureus. However, the exact causal microorganism can only be definitively identified through further diagnostic tests such as sputum culture or blood culture. The development of HAP in a patient already hospitalized for a stroke can complicate recovery and increase the risk of severe complications, including respiratory failure, sepsis, and even death. Therefore, prompt diagnosis and appropriate treatment are crucial. Despite the patient's leukocyte count being within normal limits, this does not rule out a bacterial infection, as the patient's symptoms and the findings on physical examination and chest radiography are more indicative of the diagnosis. 

Vote History: Neurology": "yes", "Infectious Disease": "yes", "Pulmonology": "yes", "Radiology": "yes", "Hematology": "yes"
\end{quote}

\begin{quote}

Question: A 2-year-old boy is brought to the physician for evaluation of pallor and increasing lethargy for 2 days. One week ago, he experienced abdominal pain, vomiting, and bloody diarrhea that have since subsided. The patient's father states that they returned early from a 6-week roadtrip in Mexico because of these symptoms. His parents have been giving him oral rehydration solution. His immunizations are up-to-date. He appears pale. His temperature is 38.4  (101.1F), pulse is 130/min, respirations are 35/min, and blood pressure is 95/50 mm Hg. Examination shows scleral icterus. The abdomen is soft and nontender; there is no rebound or guarding. Bowel sounds are hyperactive. The remainder of the examination shows no abnormalities. Laboratory studies show:  Hemoglobin 8.5 g/dL Mean corpuscular volume 94 03bcm3 Leukocyte count 18,000/mm3 Platelet count 45,000/mm3 Prothrombin time 12 sec Partial thromboplastin time 34 sec Serum Urea nitrogen 28 mg/dL Creatinine 1.6 mg/dL Bilirubin Total 2.5 mg/dL Direct 0.1 mg/dL Lactate dehydrogenase 1658 U/L A blood smear shows schistocytes. Which of the following is the most likely diagnosis?\"", "options": "A": "Henoch-Sch 00f6nlein Purpura", "B": "Thrombotic thrombocytopenic purpura", "C": "Hemolytic uremic syndrome", "D": "Immune thrombocytopenic purpura", "E": "Disseminated intravascular coagulation

Experts: "Pediatrics", "Infectious Diseases", "Hematology", "Gastroenterology", "Nephrology"

Pediatrics Expert: The medical scenario described suggests that the 2-year-old boy is suffering from Hemolytic Uremic Syndrome (HUS). HUS is a condition characterized by hemolytic anemia, acute kidney injury, and thrombocytopenia. It is most commonly caused by an infection with Shiga toxin-producing bacteria, such as Escherichia coli, which is often contracted through contaminated food or water. The recent travel to Mexico and the symptoms of abdominal pain, vomiting, and bloody diarrhea suggest a possible exposure to such bacteria. The most alarming aspects of this case are the boy's pallor and lethargy, which indicate severe anemia and possible kidney damage. The laboratory results support this diagnosis. The low hemoglobin level (8.5 g/dL) and the presence of schistocytes (fragmented red blood cells) on the blood smear are indicative of hemolytic anemia. The elevated urea nitrogen (28 mg/dL) and creatinine (1.6 mg/dL) levels suggest acute kidney injury. The low platelet count (45,000/mm3) is indicative of thrombocytopenia. The elevated lactate dehydrogenase (1658 U/L) is a marker of cell death, which is consistent with the hemolysis and kidney injury seen in HUS. The boy's high leukocyte count (18,000/mm3) suggests an ongoing inflammatory response, likely due to the underlying infection.."

Infectious Disease Expert: "The medical scenario described suggests that the 2-year-old boy is suffering from Hemolytic Uremic Syndrome (HUS). HUS is a condition characterized by hemolytic anemia, acute kidney injury, and thrombocytopenia. It is most commonly caused by an infection with Shiga toxin-producing bacteria, such as Escherichia coli, which is often contracted through contaminated food or water. The recent travel to Mexico and the symptoms of abdominal pain, vomiting, and bloody diarrhea suggest a possible exposure to such bacteria. The most alarming aspects of this case include the boy's pallor and lethargy, which are signs of anemia and possibly reduced kidney function. The laboratory results support this diagnosis: the low hemoglobin level indicates anemia, the presence of schistocytes (fragmented red blood cells) on the blood smear is indicative of hemolytic anemia, and the elevated urea nitrogen and creatinine levels suggest kidney damage. The low platelet count is also concerning as it indicates a risk of bleeding. The elevated lactate dehydrogenase level is a marker of cell damage, which is consistent with the hemolysis and kidney injury seen in HUS. The boy's elevated temperature and leukocyte count suggest an ongoing infection or inflammation.."

Hematology Expert: "The medical scenario described suggests that the child is suffering from Hemolytic Uremic Syndrome (HUS). HUS is a condition characterized by a triad of microangiopathic hemolytic anemia, thrombocytopenia, and acute kidney injury. It is most commonly caused by an infection with Shiga toxin-producing bacteria, such as Escherichia coli O157:H7, which is often contracted through contaminated food or water. The recent travel to Mexico and the episode of bloody diarrhea suggest a possible exposure to such bacteria. The most alarming aspects of this case are the laboratory findings, which are consistent with HUS. The child's hemoglobin level is low (8.5 g/dL), indicating anemia, and the blood smear shows schistocytes, which are fragmented red blood cells typically seen in microangiopathic hemolytic anemia. The platelet count is also low (45,000/mm3), indicating thrombocytopenia. The elevated lactate dehydrogenase (LDH) level (1658 U/L) is a marker of cell destruction, which is consistent with hemolysis. The elevated urea nitrogen (28 mg/dL) and creatinine (1.6 mg/dL) levels indicate kidney dysfunction, which is the third component of the HUS triad.."

Gastroenterology Expert: "The medical scenario described suggests that the 2-year-old boy is suffering from Hemolytic Uremic Syndrome (HUS). HUS is a condition characterized by hemolytic anemia, acute kidney injury, and thrombocytopenia. It is most commonly caused by an infection with E. coli O157:H7, which produces Shiga toxin. This toxin can cause damage to the endothelial cells lining the blood vessels, leading to platelet activation and aggregation, hemolysis, and kidney injury. The boy's recent history of abdominal pain, vomiting, and bloody diarrhea following a trip to Mexico suggests a possible infection with E. coli O157:H7. The pallor and lethargy are likely due to anemia and kidney injury. The laboratory findings support this diagnosis. The low hemoglobin level and presence of schistocytes (fragmented red blood cells) on the blood smear indicate hemolytic anemia. The elevated urea nitrogen and creatinine levels suggest kidney injury. The low platelet count is indicative of thrombocytopenia. The most alarming aspects of this case are the severe anemia, kidney injury, and thrombocytopenia. These can lead to serious complications, including heart failure, severe kidney damage requiring dialysis, and bleeding problems. The boy's elevated temperature also suggests that he may still have an ongoing infection. Prompt treatment is crucial to prevent further complications and improve the prognosis."

Mephrology Expert: "The medical scenario described suggests that the child is suffering from Hemolytic Uremic Syndrome (HUS). HUS is a condition characterized by a triad of microangiopathic hemolytic anemia, thrombocytopenia, and acute kidney injury. It is most commonly caused by an infection with Shiga toxin bacteria, such as Escherichia coli, which is often contracted through contaminated food or water. The recent travel to Mexico and the episode of bloody diarrhea suggest a possible exposure to such bacteria. The most alarming aspects of this case include:1. The presence of schistocytes on the blood smear: Schistocytes are fragmented red blood cells, indicative of microangiopathic hemolytic anemia, a condition where red blood cells are destroyed as they pass through small, damaged blood vessels. 2. The low platelet count (thrombocytopenia): This suggests that platelets are being consumed in the process of clot formation within the small blood vessels. 3. The elevated urea nitrogen and creatinine levels: These are signs of acute kidney injury, another key feature of HUS. 4. The symptoms of pallor and lethargy: These are likely due to anemia and kidney dysfunction. 5. The presence of scleral icterus and elevated total bilirubin: These are signs of hemolysis, or breakdown of red blood cells. 6..", "option analyses": "Pediatric Medicine": "Option A: 'Henoch-Sch Purpura' - This condition is characterized by palpable purpura, arthritis, abdominal pain, and renal disease. The patient's symptoms and lab findings do not align with this diagnosis, making it an unlikely choice. Option B: 'Thrombotic thrombocytopenic purpura' - This condition is characterized by a pentad of symptoms: thrombocytopenia, microangiopathic hemolytic anemia, neurological symptoms, fever, and renal disease. The absence of neurological symptoms and fever in the patient's presentation makes this option less likely. Option C: 'Hemolytic uremic syndrome' - This is the most likely diagnosis given the patient's symptoms and lab findings. The patient's recent history of bloody diarrhea, pallor, lethargy, and the presence of schistocytes on the blood smear, along with the elevated urea nitrogen and creatinine levels, are all indicative of HUS. Option D: 'Immune thrombocytopenic purpura' - This condition is characterized by isolated thrombocytopenia without significant anemia or renal disease. The patient's anemia and renal disease make this option less likely.."

Hematology Expert: "Option A: 'Henoch-Sch Purpura' - This condition is characterized by palpable purpura, arthritis or arthralgia, abdominal pain, and renal disease. The patient's symptoms and lab findings do not align with this diagnosis, making it an unlikely choice. Option B: 'Thrombotic thrombocytopenic purpura' - This condition is characterized by a pentad of symptoms: thrombocytopenia, microangiopathic hemolytic anemia, neurological symptoms, fever, and renal disease. The absence of neurological symptoms and fever in the patient's presentation makes this option less likely. Option C: 'Hemolytic uremic syndrome' - This condition is characterized by a triad of microangiopathic hemolytic anemia, thrombocytopenia, and acute kidney injury. The patient's symptoms and lab findings align with this diagnosis, making it the most likely choice. Option D: 'Immune thrombocytopenic purpura' - This condition is characterized by isolated thrombocytopenia without anemia or organ injury. The patient's presentation of anemia and kidney injury makes this option less likely. Option E: 'Disseminated intravascular coagulation' - This condition is characterized by widespread activation of the coagulation cascade that results in the formation of blood clots in the small blood vessels throughout the body.."

Synthezied Report:  "Question: A 2-year-old boy is brought to the physician for evaluation of pallor and increasing lethargy for 2 days. One week ago, he experienced abdominal pain, vomiting, and bloody diarrhea that have since subsided. The patient's father states that they returned early from a 6-week roadtrip in Mexico because of these symptoms. His parents have been giving him oral rehydration solution. His immunizations are up-to-date. He appears pale. His temperature is 38.4 (101.1), pulse is 130/min, respirations are 35/min, and blood pressure is 95/50 mm Hg. Examination shows scleral icterus. The abdomen is soft and nontender; there is no rebound or guarding. Bowel sounds are hyperactive. The remainder of the examination shows no abnormalities. Laboratory studies show: 8.5 g/dL corpuscular volume 94 count 18,000/mm3 count 45,000/mm3 time 12 sec Partial thromboplastin time 34 sec Serum Urea nitrogen 28 mg/dL 1.6 mg/dL 2.5 mg/dL 0.1 mg/dL dehydrogenase 1658 U/L blood smear shows schistocytes. Which of the following is the most likely diagnosis?\"  Options: {'A': 'Henoch-Sch Purpura', 'B': 'Thrombotic thrombocytopenic purpura', 'C': 'Hemolytic uremic syndrome', 'D': 'Immune thrombocytopenic purpura', 'E': 'Disseminated intravascular coagulation"'} Total Analysis: The boy's symptoms and laboratory findings are consistent with Hemolytic Uremic Syndrome (HUS). His recent travel to Mexico and the episode of bloody diarrhea suggest a possible infection with Shiga toxin-producing bacteria, such as Escherichia coli. The presence of schistocytes on the blood smear, indicative of hemolytic anemia, the low platelet count, indicative of thrombocytopenia, and the elevated urea nitrogen and creatinine levels, indicative of kidney injury, are all key features of HUS. The boy's pallor and lethargy are likely due to the anemia and kidney dysfunction. His elevated temperature and leukocyte count suggest an ongoing infection or inflammation. The most alarming aspects of this case are the severe anemia, kidney injury, and thrombocytopenia, which can lead to serious complications if not promptly treated. The options provided in the reports further support HUS as the most likely diagnosis, as the boy's symptoms and lab findings do not align with the other conditions listed.

Vote History: Pediatrics": "yes", "Infectious Diseases": "yes", "Hematology": "yes", "Gastroenterology": "yes", "Nephrology": "yes", "Pediatric Medicine": "yes"

\end{quote}

\begin{quote}
Question: A 2-year-old boy is brought to the physician for evaluation of pallor and increasing lethargy for 2 days. One week ago, he experienced abdominal pain, vomiting, and bloody diarrhea that have since subsided. The patient's father states that they returned early from a 6-week roadtrip in Mexico because of these symptoms. His parents have been giving him oral rehydration solution. His immunizations are up-to-date. He appears pale. His temperature is 38.4\u00b0C (101.1\u00b0F), pulse is 130/min, respirations are 35/min, and blood pressure is 95/50 mm Hg. Examination shows scleral icterus. The abdomen is soft and nontender; there is no rebound or guarding. Bowel sounds are hyperactive. The remainder of the examination shows no abnormalities. Laboratory studies show: Hemoglobin 8.5 g/dL Mean corpuscular volume 94 \u03bcm3 Leukocyte count 18,000/mm3 Platelet count 45,000/mm3 Prothrombin time 12 sec Partial thromboplastin time 34 sec Serum Urea nitrogen 28 mg/dL Creatinine 1.6 mg/dL Bilirubin Total 2.5 mg/dL Direct 0.1 mg/dL Lactate dehydrogenase 1658 U/L A blood smear shows schistocytes. Which of the following is the most likely diagnosis?\"",
    "options": 
        "A": "Henoch-Sch\u00f6nlein Purpura",
        "B": "Thrombotic thrombocytopenic purpura",
        "C": "Hemolytic uremic syndrome",
        "D": "Immune thrombocytopenic purpura",
        "E": "Disseminated intravascular coagulation

Experts: "Pediatrics",
        "Hematology",
        "Nephrology",
        "Infectious Disease",
        "Gastroenterology"

Pediatrics Expert: "**Interpretation of the Condition:** The 2-year-old boy is most likely suffering from **Hemolytic Uremic Syndrome (HUS)**. HUS is a serious condition characterized by the triad of: 1. **Microangiopathic Hemolytic Anemia (MAHA):** Destruction of red blood cells due to damage in small blood vessels. 2. **Thrombocytopenia:** Low platelet count leading to bleeding tendencies.3. **Acute Renal Failure (Uremia):** Sudden loss of kidney function.**Evidence Supporting the Diagnosis:**- **Recent Gastrointestinal Illness:** The boy had abdominal pain, vomiting, and bloody diarrhea one week prior. This is significant because HUS often follows an episode of gastroenteritis caused by Shiga toxin-producing bacteria, such as **Escherichia coli O157:H7**, which is common in areas with lower sanitation standards\u2014a risk increased by recent travel to Mexico.- **Anemia:** - **Low Hemoglobin:** 8.5 g/dL (normal for age ~11-13 g/dL).  - **Normocytic Anemia:** MCV is normal (94 \u03bcm\u00b3), indicating red blood cells are being destroyed rather than produced abnormally.  - **Elevated Lactate Dehydrogenase (LDH):** 1658 U/L, suggesting cell lysis. - **Indirect Hyperbilirubinemia:** Total bilirubin elevated at 2.5 mg/dL with direct bilirubin normal at 0.1 mg/dL, indicating hemolysis.- **Schistocytes on Blood Smear:** Fragmented red blood cells confirm microangiopathic hemolytic anemia.- **Thrombocytopenia:** - **Low Platelet Count:** 45,000/mm\u00b3 (normal ~150,000-450,000/mm\u00b3), increasing bleeding risk.- **Acute Renal Failure:**  - **Elevated Blood Urea Nitrogen (BUN):** 28 mg/dL.  - **Elevated Creatinine:** 1.6 mg/dL.  - These values indicate impaired kidney function.- **Leukocytosis:** Elevated white blood cell count at 18,000/mm\u00b3, often seen in response to infection or stress.- **Vital Signs and Symptoms:**  - **Fever:** 38.4\u00b0C (101.1\u00b0F).  - **Tachycardia and Tachypnea:** Pulse 130/min, respirations 35/min.  - **Pallor and Scleral Icterus:** Visible signs of anemia and jaundice due to hemolysis.  - **Hyperactive Bowel Sounds:** May indicate ongoing gastrointestinal activity or irritation.**Alarming or Noteworthy Aspects:** 1. **Rapid Onset of Hemolytic Anemia and Thrombocytopenia:** The acute drop in hemoglobin and platelets can lead to severe anemia and bleeding complications, requiring immediate management. 2. **Acute Renal Failure:** The kidneys' reduced ability to filter waste can rapidly lead to fluid and electrolyte imbalances, hypertension, and the need for dialysis if not promptly addressed. 3. **History of Shiga Toxin Exposure:** The prior episode of bloody diarrhea suggests infection with a Shiga toxin-producing E. coli strain. This toxin can severely damage endothelial cells, particularly in the kidneys, exacerbating HUS. 4. **Elevated LDH and Presence of Schistocytes:** Signify significant ongoing hemolysis, which can lead to further complications like hyperbilirubinemia and jaundice. 5. **Age of the Patient:** Young children are more susceptible to severe manifestations of HUS and may deteriorate rapidly without prompt treatment. 6. **Potential for Neurological Complications:** Although not currently presenting neurological symptoms, HUS can involve the central nervous system, leading to seizures or altered mental status. 7. **Need for Specialized Care:** Management of HUS often requires intensive care support, including possible blood transfusions, plasma exchange, and renal replacement therapy.---**Conclusion:** The clinical picture is consistent with Hemolytic Uremic Syndrome, likely precipitated by a Shiga toxin-producing E. coli infection during travel. Immediate hospitalization and supportive care are crucial to address hemolysis, prevent further kidney damage, and monitor for potential complications.",

Hematology Expert:         "Hematology": "Based on the clinical presentation and laboratory findings, the most likely diagnosis for this 2-year-old boy is **Hemolytic-Uremic Syndrome (HUS)**.**Interpretation of the Condition:**Hemolytic-Uremic Syndrome is a condition characterized by the triad of microangiopathic hemolytic anemia, thrombocytopenia, and acute kidney injury. This syndrome is often precipitated by an infection with **Shiga toxin-producing Escherichia coli (E. coli O157:H7)**, which is commonly acquired through ingestion of undercooked meat or contaminated food and water\u2014risks that can be heightened during travel to areas with different food safety standards, such as parts of Mexico.**Key Features Supporting the Diagnosis:**1. **Recent Gastrointestinal Illness:**   - The boy experienced **abdominal pain, vomiting, and bloody diarrhea** one week prior, which is typical of an E. coli O157:H7 infection.2. **Anemia and Hemolysis:**   - **Hemoglobin of 8.5 g/dL** indicates anemia.   - **Schistocytes** on blood smear signify microangiopathic hemolytic anemia due to mechanical destruction of red blood cells in small vessels clogged with microthrombi.3. **Thrombocytopenia:**  - **Platelet count of 45,000/mm\u00b3** is significantly low, reflecting consumption of platelets in the formation of microthrombi.4. **Renal Impairment:**  - Elevated **blood urea nitrogen (28 mg/dL)** and **creatinine (1.6 mg/dL)** indicate acute kidney injury secondary to microthrombi in renal vasculature.5. **Indirect Hyperbilirubinemia:**   - Elevated **total bilirubin (2.5 mg/dL)** with normal direct bilirubin suggests increased breakdown of red blood cells.6. **Elevated Lactate Dehydrogenase (LDH):**\  - **LDH of 1658 U/L** is elevated due to cellular breakdown and hemolysis.7. **Absence of Coagulation Abnormalities:**   - **Normal PT and PTT** differentiate HUS from disseminated intravascular coagulation (DIC), where these parameters are typically prolonged.**Most Alarming and Noteworthy Aspects:**- **Acute Kidney Injury:**  - The development of renal impairment in a young child is particularly concerning. If not promptly managed, it can progress to severe renal failure requiring dialysis.- **Microangiopathic Hemolytic Anemia:**  - The rapid destruction of red blood cells can lead to severe anemia, which contributes to the patient's pallor and lethargy.- **Thrombocytopenia with Risk of Bleeding:** - Although the primary issue is platelet consumption rather than destruction, low platelet levels increase the risk of bleeding complications.- **Potential for Neurological Complications:**  - While not currently present, HUS can progress to involve the central nervous system, leading to seizures or altered mental status.- **Recent Travel History:**  - The exposure to different environmental pathogens during the trip to Mexico raises concerns about infectious etiologies that may require specific public health interventions.**Conclusion:**Immediate medical intervention is crucial for managing Hemolytic-Uremic Syndrome. The primary focus should be on supportive care, including careful monitoring of fluid and electrolyte balance, management of anemia, and monitoring renal function to address any progression towards renal failure. Early recognition and treatment are essential to improve outcomes and reduce the risk of long-term complications.",

Nephrology Expert: "**Diagnosis:** The most likely diagnosis is **Hemolytic Uremic Syndrome (HUS)**.---**Interpretation of the Condition:**Hemolytic Uremic Syndrome is a serious condition characterized by the triad of:1. **Microangiopathic Hemolytic Anemia (MAHA):**  - Evidenced by anemia (hemoglobin of 8.5 g/dL) with a normocytic mean corpuscular volume (MCV of 94 \u03bcm\u00b3).   - Presence of schistocytes on the peripheral blood smear indicates red blood cell fragmentation due to mechanical injury in the microvasculature.2. **Thrombocytopenia:**   - The platelet count is significantly decreased (45,000/mm\u00b3), suggesting platelet consumption.3. **Acute Kidney Injury (AKI):**   - Elevated blood urea nitrogen (BUN) of 28 mg/dL and creatinine of 1.6 mg/dL indicate renal impairment.   - The kidneys are affected due to microthrombi formation in the renal vasculature.**Clinical Context Supporting HUS:**- **Recent Gastrointestinal Infection:**   - The patient had abdominal pain, vomiting, and **bloody diarrhea** one week ago, which are classic antecedent symptoms of HUS. - Travel history to Mexico increases exposure risk to **Shiga toxin-producing bacteria**, such as *Escherichia coli* O157:H7.- **Signs of Hemolysis:**   - Elevated lactate dehydrogenase (LDH) level (1658 U/L) is a marker of cell breakdown. - Elevated total bilirubin (2.5 mg/dL) with predominantly indirect (unconjugated) bilirubin suggests hemolysis.  - Scleral icterus is a physical sign of hyperbilirubinemia.- **Vital Signs and Physical Examination:**  - Fever (temperature of 38.4\u00b0C or 101.1\u00b0F) and tachycardia (pulse of 130/min) may result from the underlying infection and hemolysis. - Tachypnea (respirations of 35/min) could be compensatory due to anemia. - Hyperactive bowel sounds may still be present from gastrointestinal involvement.**Laboratory Findings Excluding Other Conditions:**- **Normal Coagulation Profile:**  - Prothrombin time (PT) and partial thromboplastin time (PTT) are within normal limits, making disseminated intravascular coagulation (DIC) less likely.- **Leukocytosis:**   - Elevated leukocyte count (18,000/mm\u00b3) indicates a systemic response to infection.---**Alarming and Noteworthy Aspects:**1. **Rapid Progression of Renal Dysfunction:**  - The development of AKI in a young child is concerning for potential long-term renal impairment.   - Close monitoring of kidney function is essential to prevent complications like hypertension and chronic kidney disease.2. **Severe Anemia and Thrombocytopenia:**  - The low hemoglobin and platelet counts increase the risk of hypoxia and bleeding, respectively.   - The patient's pallor and lethargy reflect significant anemia affecting oxygen delivery to tissues.3. **Evidence of Hemolysis and High LDH:**   - High LDH levels correlate with the degree of hemolysis and tissue injury.  - Hemolysis can lead to iron overload and further renal damage due to hemoglobinuria.4. **Potential Neurological Complications:**  - Although not currently present, HUS can progress to involve neurological symptoms like seizures or altered mental status.   - Vigilance for any new neurological signs is necessary.5. **Supportive Care Requirements:**   - The child may require transfusions, dialysis, or intensive care support.  - Early intervention improves outcomes, emphasizing the urgency in management.6. **Public Health Implications:**  - Shiga toxin-producing bacterial infections have implications for food safety and public health.   - Reporting and addressing the source can prevent further cases.---**Conclusion:**The patient's history, clinical presentation, and laboratory findings are consistent with Hemolytic Uremic Syndrome secondary to a Shiga toxin-producing bacterial infection, likely acquired during travel in Mexico. Immediate medical attention is warranted to manage the hematological abnormalities and prevent further renal damage.",

Infectious Diease Expert: "Based on the clinical scenario, the most likely diagnosis for this 2-year-old boy is **Hemolytic Uremic Syndrome (HUS)**.**Analysis:**- **Recent Gastrointestinal Infection:** The patient had abdominal pain, vomiting, and bloody diarrhea one week ago, which are suggestive of a gastrointestinal infection. Traveling to Mexico increases the risk of exposure to enteropathogens, particularly **Shiga toxin-producing Escherichia coli (STEC)**, commonly E. coli O157:H7.- **Anemia with Schistocytes:** The hemoglobin level is low at 8.5 g/dL, indicating anemia. The presence of schistocytes on the blood smear suggests a microangiopathic hemolytic anemia, where red blood cells are being fragmented within small vessels.- **Thrombocytopenia:** The platelet count is low at 45,000/mm\u00b3, indicating thrombocytopenia, which is a hallmark of HUS.- **Elevated Lactate Dehydrogenase (LDH):** LDH is significantly elevated at 1658 U/L, consistent with cell breakdown due to hemolysis.- **Renal Impairment:** Elevated blood urea nitrogen (28 mg/dL) and creatinine (1.6 mg/dL) levels indicate acute kidney injury, which is a critical component of HUS.- **Indirect Hyperbilirubinemia:** The total bilirubin is elevated (2.5 mg/dL) with a normal direct bilirubin, suggesting indirect (unconjugated) hyperbilirubinemia due to increased breakdown of red blood cells.- **Vital Signs and Physical Examination:** The patient is febrile (38.4\u00b0C), tachycardic (130/min), and tachypneic (35/min). Pallor and scleral icterus further support anemia and jaundice due to hemolysis. Hyperactive bowel sounds may reflect ongoing gastrointestinal hyperactivity.**Noteworthy and Alarming Aspects:**- **Progression to Acute Kidney Injury:** The development of renal impairment in a young child is particularly concerning. HUS can lead to significant kidney damage, and prompt recognition is essential to prevent long-term renal complications.- **Microangiopathic Hemolytic Anemia and Thrombocytopenia:** The combination of hemolytic anemia with schistocytes and low platelet count indicates a serious systemic process affecting the microvasculature.- **Recent Exposure to Pathogenic Bacteria:** The history of bloody diarrhea following travel to Mexico suggests exposure to STEC, which produces Shiga toxin leading to endothelial damage and the subsequent triad of HUS.- **Risk of Neurological Complications:** While not currently present, HUS can progress to involve neurological symptoms such as seizures or altered mental status, which requires close monitoring.- **Need for Immediate Medical Intervention:** Early diagnosis and management are crucial. Supportive care, including hydration and monitoring of renal function, is essential. In severe cases, dialysis may be required.**Conclusion:**The patient's presentation aligns with Hemolytic Uremic Syndrome secondary to a Shiga toxin-producing E. coli infection. Immediate medical intervention is necessary to manage the acute symptoms and prevent potential complications.",

Gastroenterology Expert: "**Diagnosis: Hemolytic Uremic Syndrome (HUS)**---**Interpretation of the Condition:**The 2-year-old boy is exhibiting classic signs and laboratory findings consistent with Hemolytic Uremic Syndrome (HUS), a condition characterized by the triad of microangiopathic hemolytic anemia, thrombocytopenia, and acute kidney injury. HUS is a serious complication that often follows an episode of gastroenteritis caused by Shiga toxin-producing bacteria, such as *Escherichia coli* O157:H7.**Key Features Supporting the Diagnosis:**1. **Recent Gastrointestinal Infection:**   - *Bloody Diarrhea History:* One week prior, the patient experienced abdominal pain, vomiting, and bloody diarrhea. This is indicative of an acute gastrointestinal infection, likely caused by Shiga toxin-producing *E. coli* acquired during travel in Mexico. - *Resolution of GI Symptoms:* The gastrointestinal symptoms have since subsided, which is a common progression before the onset of HUS.2. **Microangiopathic Hemolytic Anemia:**   - *Anemia and Pallor:* Hemoglobin is low at 8.5 g/dL, and the patient appears pale.   - *Schistocytes on Blood Smear:* The presence of fragmented red blood cells indicates mechanical destruction within the microvasculature.  - *Elevated Lactate Dehydrogenase (LDH):* LDH is significantly elevated at 1658 U/L, reflecting ongoing hemolysis.  - *Indirect Hyperbilirubinemia:* Total bilirubin is elevated with a predominantly indirect (unconjugated) component, consistent with hemolysis.  - *Scleral Icterus:* Physical examination shows jaundice of the sclera due to increased bilirubin.3. **Thrombocytopenia:**   - *Low Platelet Count:* Platelets are decreased at 45,000/mm\u00b3, indicating consumption within thrombi in small vessels.   - *Normal Coagulation Studies:* Prothrombin time (12 sec) and partial thromboplastin time (34 sec) are within normal limits, suggesting that the thrombocytopenia is not due to a coagulation disorder like disseminated intravascular coagulation (DIC).4. **Acute Kidney Injury:**  - *Elevated BUN and Creatinine:* Serum urea nitrogen is elevated at 28 mg/dL, and creatinine is elevated at 1.6 mg/dL, indicating renal dysfunction.   - *Oliguria or Anuria (if present):* While not specified, patients with HUS often have decreased urine output.5. **Additional Findings:**  - *Vital Signs Indicate Tachycardia and Tachypnea:* Pulse of 130/min and respirations of 35/min may reflect compensation for anemia and metabolic acidosis from renal failure.  - *Leukocytosis:* Elevated leukocyte count at 18,000/mm\u00b3 may be a response to infection or stress.**Most Alarming or Noteworthy Aspects:**1. **Rapid Progression to Renal Failure:**  - The development of acute kidney injury in a young child is particularly concerning. Without prompt management, this can progress to severe renal failure requiring dialysis.2. **Hemolytic Anemia and Risk of Ischemia:**  - Ongoing hemolysis can lead to significant anemia, reducing oxygen delivery to vital organs and potentially causing organ dysfunction.3. **Thrombocytopenia and Bleeding Risk:**  - Although coagulation parameters are normal, low platelet counts increase the risk of bleeding, which can be exacerbated in vital organs.4. **Potential Neurological Complications:**   - HUS can sometimes involve neurological symptoms such as seizures or alterations in consciousness due to microthrombi affecting cerebral circulation.5. **Dehydration and Electrolyte Imbalances:**  - The initial vomiting and diarrhea may have led to dehydration, and renal dysfunction can cause dangerous electrolyte abnormalities.6. **Recent Travel History:**  - Exposure to pathogens endemic in Mexico increases the suspicion of a Shiga toxin-producing bacterial infection.**Conclusion:**The patient's clinical presentation and laboratory findings are most consistent with Hemolytic Uremic Syndrome, likely triggered by a Shiga toxin-producing *E. coli* infection contracted during travel. Immediate medical attention is required to manage renal failure, support hemoglobin levels, monitor for neurological complications, and prevent further hemolysis and thrombocytopenia.---**Recommendations:**- **Hospital Admission:** For close monitoring and management- **Renal Support:** Potential initiation of dialysis if indicated.- **Fluid and Electrolyte Management:** Careful correction of imbalances.- **Blood Transfusions:** If anemia becomes severe or symptomatic.- **Avoid Antibiotics and Anti-Motility Agents:** These can worsen the release of Shiga toxin.- **Monitoring for Complications:** Regular neurological assessments and laboratory monitoring.**Note:** Early recognition and management of HUS are crucial to improve outcomes and reduce the risk of long-term renal impairment."

Synthesized Report: "Question: A 2-year-old boy is brought to the physician for evaluation of pallor and increasing lethargy for 2 days. One week ago, he experienced abdominal pain, vomiting, and bloody diarrhea that have since subsided. The patient's father states that they returned early from a 6-week roadtrip in Mexico because of these symptoms. His parents have been giving him oral rehydration solution. His immunizations are up-to-date. He appears pale. His temperature is 38.4\u00b0C (101.1\u00b0F), pulse is 130/min, respirations are 35/min, and blood pressure is 95/50 mm Hg. Examination shows scleral icterus. The abdomen is soft and nontender; there is no rebound or guarding. Bowel sounds are hyperactive. The remainder of the examination shows no abnormalities. Laboratory studies show:Hemoglobin 8.5 g/dLMean corpuscular volume 94 \u03bcm3Leukocyte count 18,000/mm3Platelet count 45,000/mm3Prothrombin time 12 secPartial thromboplastin time 34 secSerumUrea nitrogen 28 mg/dLCreatinine 1.6 mg/dLBilirubinTotal 2.5 mg/dLDirect 0.1 mg/dLLactate dehydrogenase 1658 U/LA blood smear shows schistocytes. Which of the following is the most likely diagnosis?\" Options: {'A': 'Henoch-Sch\u00f6nlein Purpura', 'B': 'Thrombotic thrombocytopenic purpura', 'C': 'Hemolytic uremic syndrome', 'D': 'Immune thrombocytopenic purpura', 'E': 'Disseminated intravascular coagulation\\"'} Total Analysis: **The patient presents with the classic features of **Hemolytic Uremic Syndrome (HUS)**, a condition most commonly seen in young children and often triggered by an infection with Shiga toxin-producing *Escherichia coli* (EHEC), such as E. coli O157:H7. The recent travel to Mexico increases the likelihood of exposure to this pathogen through contaminated food or water.**Supporting Evidence for HUS:**1. **History of Gastrointestinal Infection:**   - The preceding episode of abdominal pain, vomiting, and bloody diarrhea is typical of EHEC infection, which produces Shiga toxin leading to endothelial damage.2. **Microangiopathic Hemolytic Anemia (MAHA):**   - Low hemoglobin (8.5 g/dL) with normocytic MCV indicates anemia due to red blood cell destruction rather than decreased production.   - Schistocytes (fragmented red blood cells) on the blood smear confirm MAHA.  - Elevated LDH and indirect hyperbilirubinemia are consistent with hemolysis.   - Physical signs like pallor and scleral icterus support the presence of anemia and jaundice from hemolysis.3. **Thrombocytopenia:**   - Significantly reduced platelet count (45,000/mm\u00b3) suggests platelet consumption in microthrombi within small vessels.. **Acute Kidney Injury:**   - Elevated BUN and creatinine levels indicate renal impairment due to microthrombi obstructing the renal vasculature.  - HUS is known to cause acute kidney injury in children, potentially progressing to renal failure if not promptly managed.5. **Leukocytosis:**  - Elevated white blood cell count may reflect systemic inflammatory response to infection or stress.6. **Normal Coagulation Studies:**   - Normal PT and PTT help differentiate HUS from disseminated intravascular coagulation (DIC), where coagulation times are typically prolonged.**Exclusion of Other Diagnoses:**- **Thrombotic Thrombocytopenic Purpura (TTP):**  - Less likely due to the patient's age (TTP is rare in young children) and absence of neurological symptoms.- **Immune Thrombocytopenic Purpura (ITP):**  - Unlikely because ITP presents with isolated thrombocytopenia without anemia, hemolysis, or renal impairment.- **Henoch-Sch\u00f6nlein Purpura (HSP):**  - Unlikely due to the absence of palpable purpura, arthralgias, and typical renal findings of hematuria and proteinuria rather than acute kidney injury.**Management and Recommendations:**- **Immediate Hospitalization:  - Close monitoring and supportive care are essential to address hemolysis, manage anemia and thrombocytopenia, and monitor renal function.- **Supportive Care:*  - Careful fluid and electrolyte management to maintain renal perfusion without overloading the compromised kidneys. - Possible blood transfusions if anemia is severe or symptomatic.  - Dialysis may be necessary if renal function continues to decline.- **Avoid Antibiotics and Anti-motility Agents:**  - Antibiotics can increase the release of Shiga toxin from bacterial lysis, potentially worsening the condition.  - Anti-motility drugs may prolong toxin exposure in the intestines.- **Monitoring for Complications:**  - Regular neurological assessments to detect any central nervous system involvement.  - Continuous monitoring of blood counts, renal function, and electrolytes.**Conclusion:**The collective clinical findings and laboratory results point toward Hemolytic Uremic Syndrome as the most likely diagnosis. Prompt recognition and management are crucial to improve outcomes and reduce the risk of long-term complications such as chronic kidney disease. Early intervention with supportive care can significantly enhance the prognosis for this patient.",

Vote History:"Pediatrics": "yes",
            "Hematology": "yes",
            "Nephrology": "yes",
            "Infectious Disease": "yes",
            "Gastroenterology": "yes"

\end{quote}

\end{document}